\documentclass[a4paper,fleqn]{cas-dc}

\usepackage[numbers]{natbib}
\usepackage{moreverb,url}
\usepackage{multirow}
\usepackage{makecell}
\usepackage{threeparttable}
 
\def\tsc#1{\csdef{#1}{\textsc{\lowercase{#1}}\xspace}}
\tsc{WGM}
\tsc{QE}
\tsc{EP}
\tsc{PMS}
\tsc{BEC}
\tsc{DE}

\begin{document}
\let\WriteBookmarks\relax
\def\floatpagepagefraction{1}
\def\textpagefraction{.001}
\shorttitle{Survey Utility based Cognitive Robotics}
\shortauthors{Qin Yang}

\title [mode = title]{Utility Theory based Cognitive Modeling in the Application of Robotics: A Survey}                      
\tnotemark[1]

\tnotetext[1]{This work is supported by the NSF Foundational Research in Robotics (FRR) Award \#2348013: ``Cooperative Multi-Agent Systems Cognitive Modeling". 

\author[1]{Qin Yang}[type=editor,
                        auid=000,bioid=1,
                        prefix=Dr.,
                        orcid=0000-0001-5342-1798]
\cormark[1]
\fnmark[1]
\ead{is3rlab@gmail.com}
\ead[url]{https://www.is3rlab.org/}

\credit{Conceptualization of this study, Methodology, Software}

\affiliation[1]{organization={College of Engineering and Computer Science, The University of Tulsa},
                city={Tulsa},
                postcode={74104}, 
                state={Oklahoma},
                country={US}}

%

\credit{Data curation, Writing - Original draft preparation}

%
%

\cortext[cor1]{Corresponding author}


\begin{abstract}
Cognitive modeling, which explores the essence of cognition, including motivation, emotion, and perception, has been widely applied in the artificial intelligence (AI) agent domains, such as robotics. From the computational perspective, various cognitive functionalities have been developed through utility theory to provide a detailed and process-based understanding for specifying corresponding computational models of representations, mechanisms, and processes.
Especially for decision-making and learning in multi-agent/robot systems (MAS/MRS), a suitable cognitive model can guide agents in choosing reasonable strategies to achieve their current needs and learning to cooperate and organize their behaviors, optimizing the system's utility, building stable and reliable relationships, and guaranteeing each group member's sustainable development, similar to the human society.
This survey examines existing robotic systems for developmental cognitive models in the context of utility theory. We discuss the evolution of cognitive modeling in robotics from behavior-based robotics (BBR) and cognitive architectures to the properties of value systems in robots, such as the studies on motivations as artificial value systems, and the utility theory based cognitive modeling for generating and updating strategies in robotic interactions. Then, we examine the extent to which existing value systems support the application of robotics from an AI agent cognitive modeling perspective, including single-agent and multi-agent systems, trust among agents, and human-robot interaction. Finally, we survey the existing literature of current value systems in relevant fields and propose several promising research directions, along with some open problems that we deem necessary for further investigation.
\end{abstract}

\begin{graphicalabstract}
\end{graphicalabstract}

\begin{highlights}
\item Robots' Motivation Mechanism Definition and Modeling
\item Robots' Value System Cognitive Modeling through Utility Theory
\item Artifical Social System integrating with Human Society
\end{highlights}

\begin{keywords}
Cognition \sep Utility \sep Needs \sep Motivation \sep Value Systems
\end{keywords}

\maketitle

\section{Introduction}

When people study, analyze, and design artificial intelligence (AI) agent or robotic systems, they would like to compare them with natural systems working. For a living organism, cognition refers to its capacity to process perceptual information and thereby manipulate its behaviors, which reflects the level of complexity and intelligence of the agent \cite{merrick2017value}.
For example, many natural systems (e.g., brains, immune system, ecology, societies) are characterized by apparently complex behaviors that emerge as a result of often nonlinear spatiotemporal interactions among a large number of component systems at different levels of the organization \cite{levin1998ecosystems}. Moreover, the cognitive capabilities of humans are generally regarded as the manifestation of self-awareness, perception, learning, knowledge, reasoning, planning, and decision-making \cite{begum2009computational}. The related research primarily focused on enhancing the capabilities of artificial systems and deepening our understanding of the fundamental roots of intelligence in natural systems \cite{lieto2016human,weng2001autonomous}, such as artificial development \cite{sandini1997human}, epigenetic robotics \cite{morse2010epigenetic}, development robotics \cite{cangelosi2015developmental,lungarella2003developmental}, and cognitive developmental robotics (CDR) \cite{asada2009cognitive}.
Specifically, artificial or mental development refers to a brain-like natural or artificial system that improves its capabilities under the control of its species-specific developmental program \cite{weng2001autonomous}. Developmental and epigenetic robotics studies the mechanisms, architectures, and constraints that allow lifelong and open-ended learning of new skills and knowledge in embodied machines \cite{lungarella2003developmental,morse2010epigenetic}. Furthermore, cognitive developmental robotics aims to provide a new understanding of how human higher cognitive functions develop by using a synthetic approach that developmentally constructs cognitive functions \cite{asada2001cognitive}. 
\begin{figure*}
	\centering
    \includegraphics[width=\linewidth]{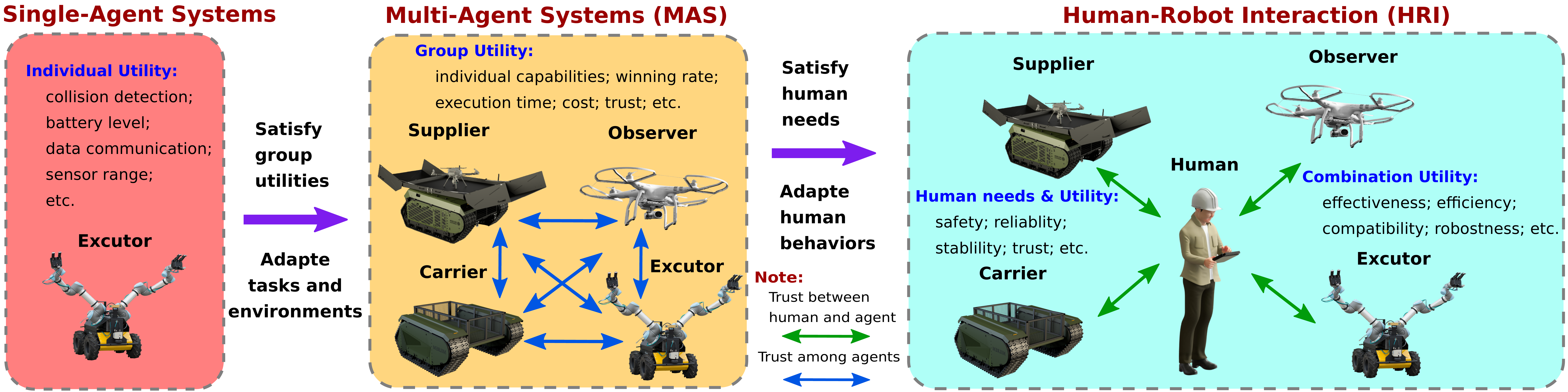}
    \caption{Illustration of the utility-orient needs paradigm based on agents' motivations among single-agent, MAS, and HRI.}
    \label{fig:overview}
\end{figure*}

The history of cognitive modeling in robotics can be traced back to behavior-based robotics (BBR) \cite{arkin1998behavior}, which is based on a biologically inspired philosophy supporting parallel, decentralized architectures and allowing some freedom of interpretation. BBR emphasizes the ``brains" of the robots rather than their ``bodies", which moves the details of robot sensors and actuators firmly into the background. It tries to understand what behaviors we should expect robots to exhibit and which computational mechanisms can achieve them. In behavior-based systems, the robot controller consists of a collection of behaviors, each of which achieves and/or maintains a specific goal, such as avoiding collision, finding targets, and wandering \cite{mataric2018integration}. In the multi-robot systems (MRS) and swarm robotics domain, behavior-based modeling focuses on developing generic mechanisms that result in coordinated group behavior, either implicitly or explicitly \cite{dorigo2002genetics}, such as formation control \cite{balch2002behavior} and cooperative architecture design \cite{parker1995design}. However, BBR mainly focuses on simple behaviors and rule-based design, and it is hard to achieve complex behaviors and build sophisticated relationships among agents (like trust, respect, and ethic), especially in human-robot interaction (HRI).

With the development of AI and cognitive science, cognitive architectures are increasingly incorporated into autonomous robot design, giving rise to new disciplines, such as cognitive robotics \cite{cangelosi2022cognitive}, and promoting the application of cognitive modeling in robotics. Specifically, cognitive modeling deals with implanting human-like cognition, albeit reduced complexity, in autonomous robots with the hope that the cognitive capacities, as they work in humans and other biological creatures, will also assist the new generation of robots to be coherent, self-motivated, social, persistent in behavior, and aware of themselves and their environments \cite{begum2009computational}. It combines the neuroanatomy, cognitive neuroscience, neuroeconomics, cognitive psychology, developmental studies, linguistics, and AI with robotics. The majority of research on cognitive modeling in robots focuses on developing models of discrete cognitive abilities, such as attention, learning, value systems, reasoning, and decision-making. The majority of research on cognitive modeling in robots focuses on developing models of discrete cognitive abilities, such as attention, learning, value systems, reasoning, and decision-making. In particular, studies on developing clear metrics and benchmarks for the different aspects of HRI aim to endow robots with advanced cognitive and communicative abilities \cite{aly2017metrics}.

From multi-agent systems (MAS) perspective, cognitive models can provide a substantial way to better probe multi-agent issues by taking into account the essential characteristics of cognitive agents and their various capacities \cite{sun2001implicit}. Furthermore, cognitive modeling presents a more realistic basis for understanding multi-agent interaction by embodying realistic constraints (like agents' innate-values and needs), capabilities, and tendencies of individual agents in interacting with their environments, including physical and social environments \cite{sun2001cognitive}. Agents potentially share the properties of {\it swarm intelligence} \cite{beni1993swarm} in practical applications (like search, rescue, mining, map construction, exploration, etc.), representing the collective behavior of distributed and self-organized systems. There are a lot of research fields related to it, especially building so-called \textit{artificial social systems}, such as drones and self-driving cars. Specifically, in multi-robot systems (MRS), each robot can internally estimate the utility of executing an action. And robots' utility estimates will be inexact for several reasons, including mission cost, strategy rewards, sensor noise, general uncertainty, and environmental change \cite{gerkey2003multi,yang2019self}. 
Furthermore, in MAS cooperation, trust is an essential component for describing the agent's purposes. In other words, there are complex relationships between utility, transparency, and trust, which depend on individual purposes \cite{wortham2017robot}. Especially considering humans and robots collaborated as a team, recognizing the utility of robots will increase interest in group understanding and improve Human-Robot Interaction (HRI) performance in tasks \cite{prewett2010managing}. 


However, bringing all of these capabilities, such as self-awareness, perception, knowledge, reasoning, planning, decision-making, and learning, together in a single artificial agent (a robot) remains one of the grand challenges, let alone multi-agent systems and human-robot interaction. 
One of the significant challenges in designing AI and robotics systems is how to formalize and quantify motivations \cite{weng1999developmental,weng2004developmental,asada2009cognitive,thrun1995lifelong}. The problem of producing structures that address the compositional construction of useful knowledge representations, relating the innate drives and needs of an intelligent system to the characteristics of the environment in which it finds itself, is of paramount importance \cite{romero2019simplifying}.
Another challenges is how to appropriately describe the relationships of their interconnected elements and natures.
Related issues include developing coordination strategies (that enable groups of agents to solve problems together effectively), negotiation mechanisms (that facilitate the interaction of a set of agents), conflict detection and resolution strategies, and other mechanisms whereby agents can contribute to overall system effectiveness while maintaining autonomy.

Generally speaking, establishing goal/sub-goal hierarchies in agents' motivational systems for open-ended learning situations and modeling their value systems based on utility theory is an open research problem, especially when their state spaces are continuous and ambiguous \cite{romero2018utility,baldassarre2014intrinsic}. Furthermore, in social scenarios, the interaction dynamics are typically unknown and unpredictable, so robots working in these environments must have appropriate decision-making capabilities to select their actions and successfully fulfill the intended tasks autonomously \cite{maroto2023systematic}.
Fig. \ref{fig:overview} outlines the vital relationships of the utility-orient needs paradigm based on agents' motivations among single-agent, MAS, and HRI, which affect systems' abilities to meet the requirements of specific tasks and adapt to human needs and uncertain environments. It presents how the utility-orient relationships affect the performance of different systems, especially evaluating the trust level among agents through various types of utilities and integrating human needs and agents' utilities to build a harmonious team \cite{yang2022self}.
\begin{figure}
	\centering
    \includegraphics[width=\columnwidth]{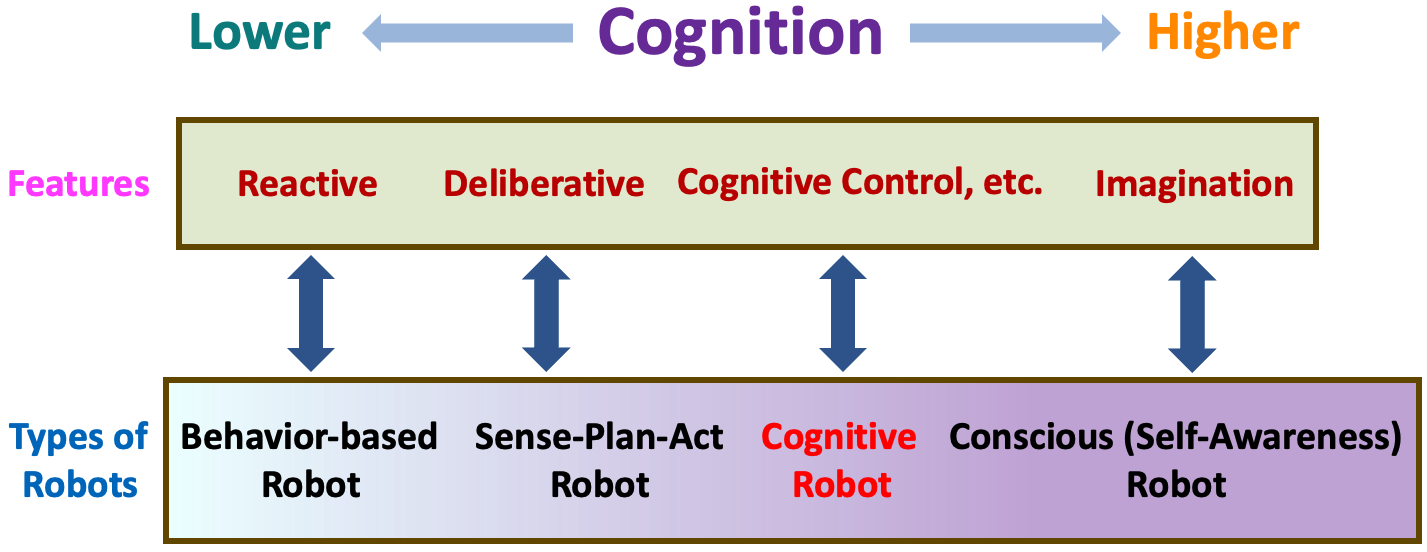}
   \caption{The illustration of relationships between the cognition evolution and robot development \cite{kawamura2009cognitive}.}
    \label{fig:cr}
\end{figure}

Following the classical expectancy-value theory \cite{wigfield2000expectancy}, the utility theory is regarded as a paradigm for describing the goals and motivations of an AI agent. A utility model serves as a function that maps the utility to a goal for each point in the state space, allowing goals to be reached by following paths of increasing utility within the state space.
Previously, K. Merrick \cite{merrick2017value} discussed the value systems for developmental cognitive robotics. More recently, M. Maroto-G´omez \cite{maroto2023systematic} provided a comprehensive overview of control architectures for autonomous and social robots.
However, a deeper analysis of the current state of utility-based cognitive modeling in robotics, including assessing its evolution over the years and framing its challenges and future goals, is necessary. Therefore, we propose this contribution, which fills a gap in the literature by providing a survey of utility theory based cognitive modeling in robotics.

The remainder of this paper is organized as follows: 
Section \ref{foundations} presents the foundations of cognitive modeling in robotics, including BBR, cognitive architectures, and utility-based models in cognition. Sections \ref{sas} and \ref{mas} review the utility theory based cognitive modeling of single-agent and multi-agent systems in the theoretical foundation, fundamental AI architectures, and autonomous and social robots by area of application. Section \ref{taa} analyzes the categories of trust in robotics based on cognitive modeling and surveys the impressive body of work from recent years. Section \ref{hri} examines how utility-based value systems integrate into human-robot interaction and have been evaluated by the research community. Section \ref{iac} proposes several open questions for future research in related fields. We conclude in Section \ref{con} with a discussion of our survey and provide our own experience as researchers.

\section{Foundations of Cognitive Modeling in Robotics}
\label{foundations}

Cognitive modeling in robots involves multiply disciplines, such as robotics, AI, computer science, cognitive science, neuroscience, neuroeconomics \cite{glimcher2013neuroeconomics}, biology, control theory, decision making theory \cite{fishburn1970utility,mongin1998expected}, and psychology. 
This section presents the methodology related to the evolution of cognitive modeling from traditional behavior-based modeling and cognitive architectures to utilizing the concept of utility to improve cognitive robots' performance. This evolution can be classified into two main stages. Behavior-based robotics characterize the first stage as extensions of theories from various fields. The second stage corresponds to cognitive robotics that focus on optimizing systems' capabilities by unifying different concepts based on utility theory.
Fig. \ref{fig:cr} illustrates relationships between cognition evolution and robot development. 

\subsection{Behavior-based Robotics}
\begin{figure}
	\centering
    \includegraphics[width=\linewidth]{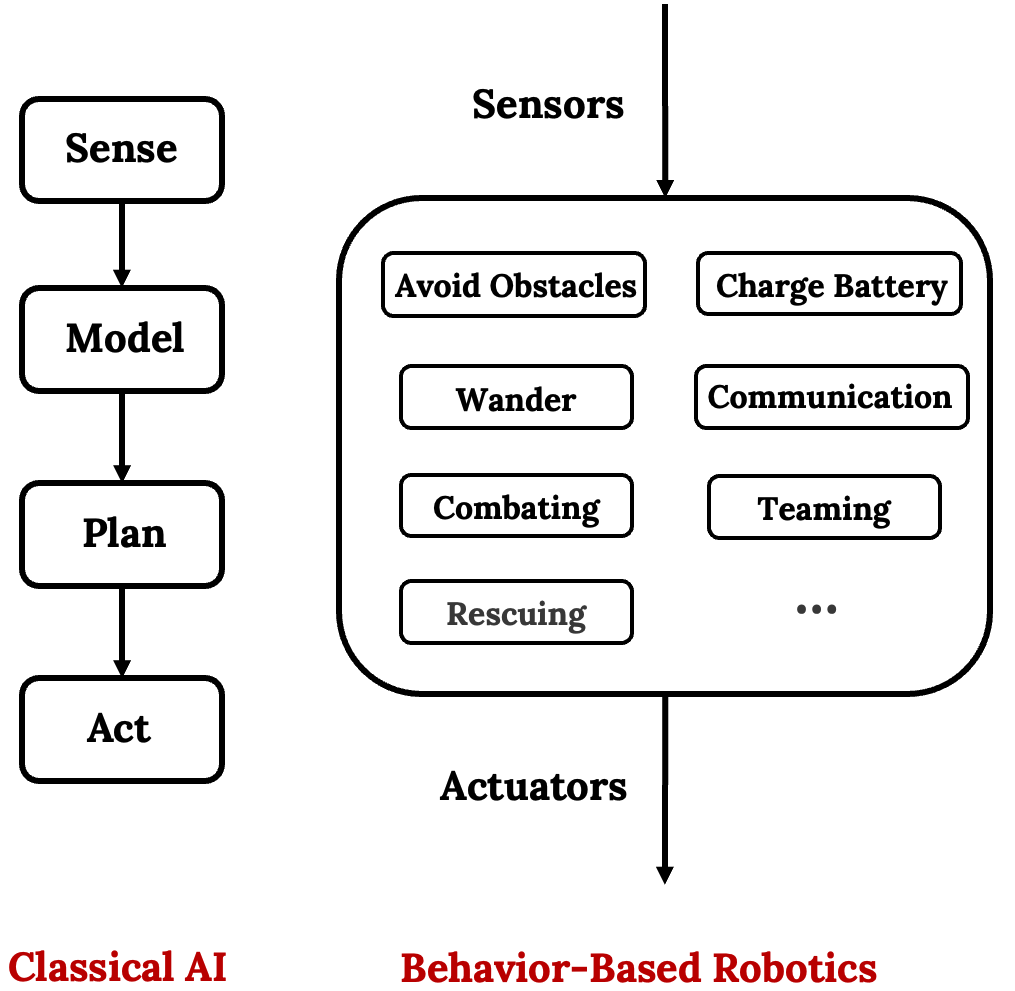}
    \caption{The illustration of the information flow in classical AI (left panel) and in BBR
(right panel).}
    \label{fig:ai_brr}
\end{figure}

\subsubsection{Features of Behavior-based Robots}
To achieve human-level intelligence, researchers have been working for about a century. Many efforts in robotics research have been inspired by simple biological organisms, with the aim of understanding and implementing basic survival-related behaviors in robots before progressing to more advanced behaviors involving, for example, high-level reasoning \cite{wahde2012introduction}. In the classical AI framework, the robot perceives the environment through sensors and builds a world model to predict the effects of various actions within it, enabling reasoning before executing behaviors in the real world. However, the procedure is very different from the distributed computation of the biological brains, and it is not the function of most biological organisms. Furthermore, Rodney Brooks argued that building world models and reasoning using explicit symbolic representational knowledge at best was an impediment to timely robotic response and at worst actually led robotics researchers in the wrong direction \cite{brooks1999cambrian}. 

Distinct from traditional AI, behavior-based robotics (BRR) provides a paradigm for exhibiting complex behaviors with little internal state, modeling its immediate environment for adaptation, and gradually correcting robots’ actions via sensory-motor links \cite{arkin1998behavior}. In BBR, intelligent behavior can be defined as the ability to survive and to strive to reach other goals in an unstructured environment\footnote{An unstructured environment is one that changes rapidly and unexpectedly, making it impossible to rely on predefined maps, which holds true for most real-world environments.}. 
Fig. \ref{fig:ai_brr} illustrates the information flow of classical AI and the intelligent behaviors built into BRR in a bottom-up fashion, starting with simple behaviors executed simultaneously in a given robotic brain and providing suggestions for which actions the robot should take. 

BBR is a branch of robotics that bridges AI, engineering, and cognitive science, providing approaches to embedded system control across research and practical applications. It aims to develop methods for controlling artificial systems, ranging from physical robots to simulated ones and other autonomous software agents, and to use robotics to model and understand biological systems more fully, typically, animals ranging from insects to humans \cite{mataric1998behavior}.
Behavior-based systems employ a collection of concurrently executing behaviors, processes connecting between sensors and effectors. An important property of behavior-based systems is its ability to contain state, construct, and utilize distributed representations simultaneously \cite{nicolescu2002hierarchical}.

\subsubsection{System Organization}
\begin{figure}
	\centering
    \includegraphics[width=\linewidth]{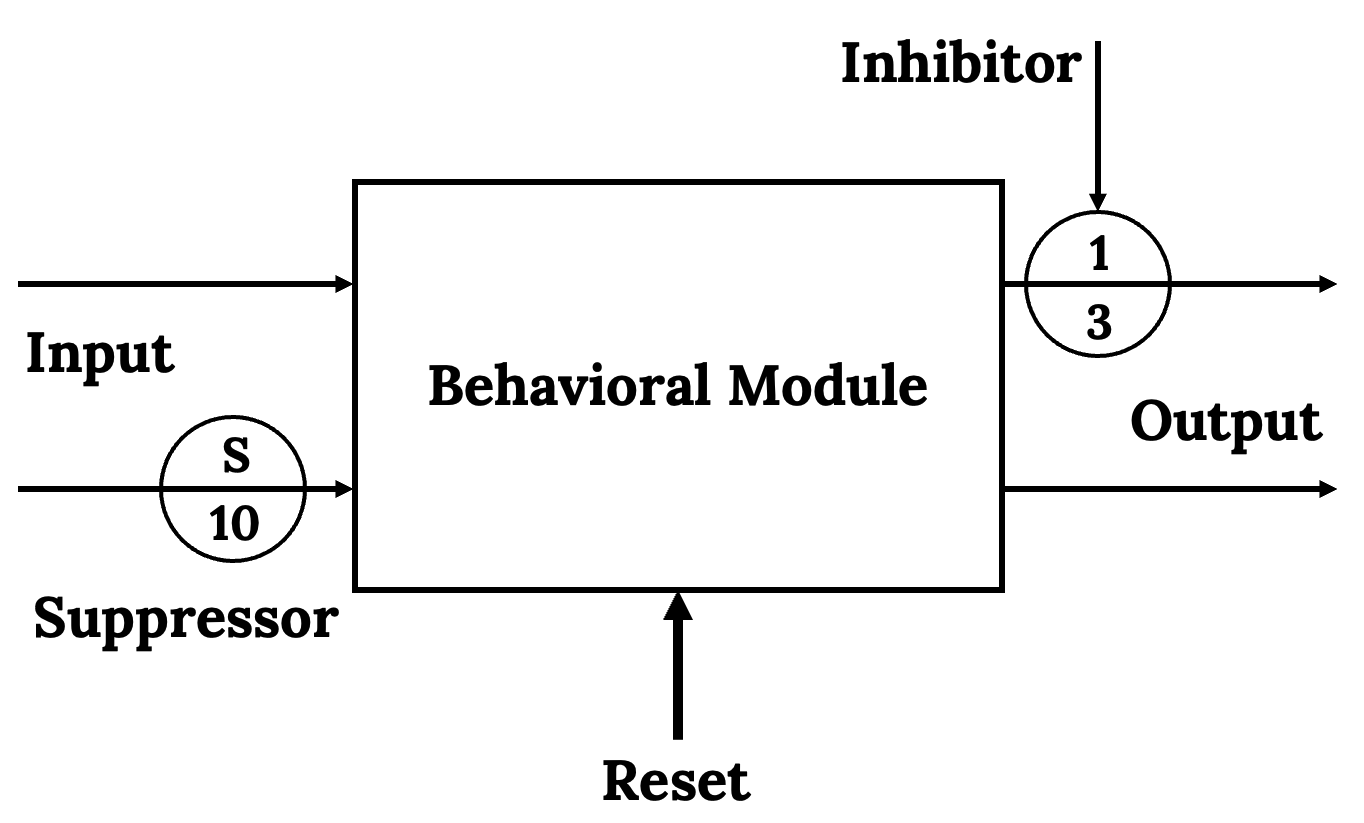}
    \caption{The illustration of the behavioral module in BRR.}
    \label{fig:bm}
\end{figure}

In behavior-based systems, the robot controller consists of a set of behaviors that achieve and/or maintain specific goals. Each behavior is regarded as a processing element that serves as a control law and can be implemented in software or hardware. Moreover, it can receive information from the robot’s sensors, such as cameras and ultrasound, infrared, or tactile sensors, and/or from other behaviors, and send results to the robot’s effectors, like wheels, grippers, arms, or speech, and/or to other behaviors in the system \cite{mataric1998behavior}. From a control theory perspective, a structured network of interacting behaviors governs a behavior-based robot.

Different behavior-based robots have some common basic features, such as survival and safety. As a top priority, behavior-based robots need to provide the most basic behaviors, such as obstacle avoidance and battery charging, to ensure normal function in an unstructured environment. Then, by generating several behaviors in operation simultaneously, the final action taken by the robot corresponded either to an optimal choice among the suggestions from the various behaviors or to an average action formed from the suggested actions. To achieve autonomous robot interaction without direct human supervision, BRR introduces the critical concept of {\it situatedness}: Behavior-based robots do not build complex, abstract world models but adapt to real-world situations. Their behavior is reactive, with direct coupling between sensors and actuators. Moreover, their internal states usually reflect motivations stored in short-term memory, which generate current goals or long-term achievement.
\begin{figure}
	\centering
    \includegraphics[width=\linewidth]{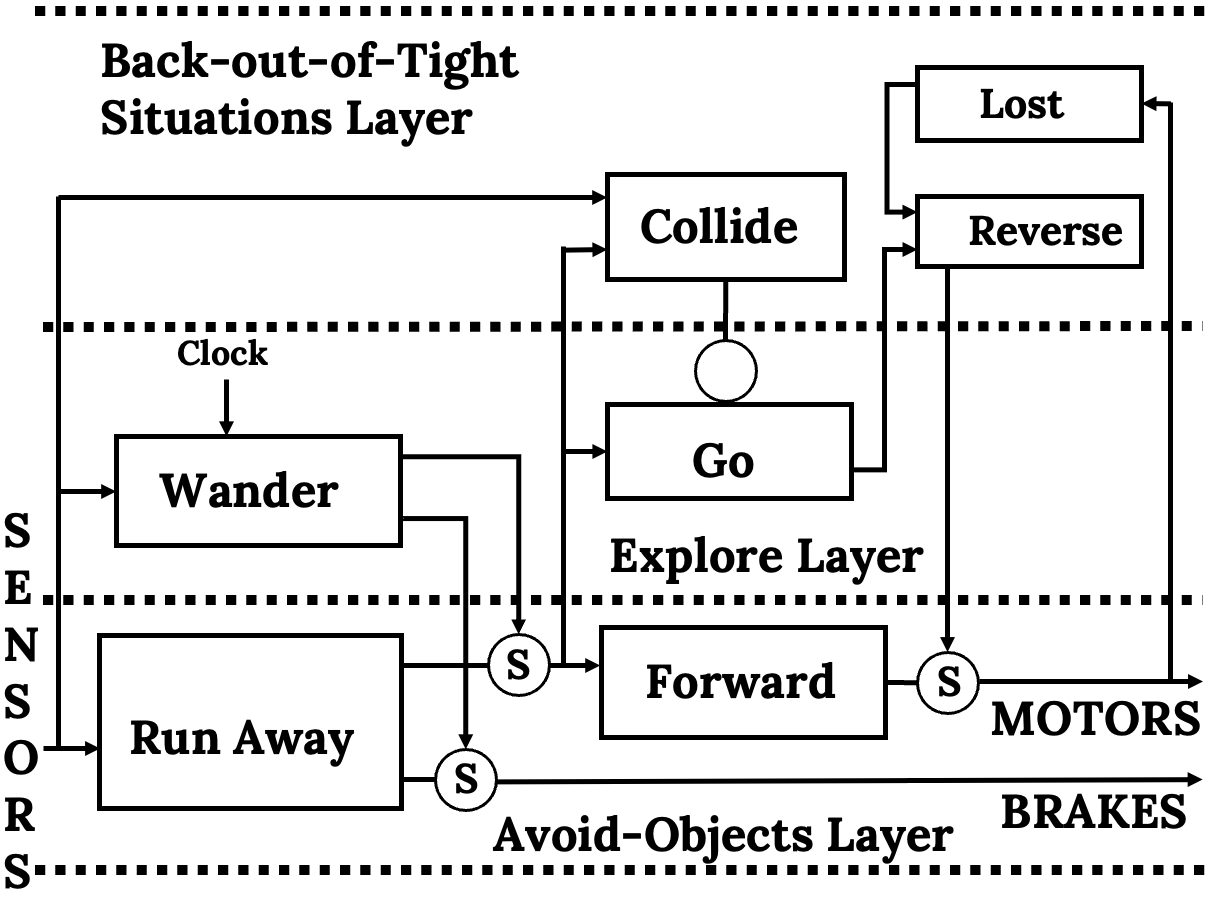}
    \caption{The illustration of the simple three-layered robot architecture \cite{brooks1999cambrian}.}
    \label{fig:bbr_layers}
\end{figure}

Generally speaking,  a set of basic behaviors is regarded as a brain organ that organizes the actions of the behavior-based robot, termed the {\it behavioral repertoire}. The process involves two steps: defining individual behaviors and selecting actions. Especially in any robot intended for complex applications, the behavioral selection system for the complex decision-making process should balance various factors and is as important as the individual behaviors themselves \cite{nicolescu2002hierarchical}.

The key challenge for behavior-based systems is coordinating multiple behaviors. To simplify the systems, most implementations use a built-in or fixed-priority ordering of behaviors. Considering computation efficiency and ease of analysis, common approaches select a behavior by computing a function of the behavior's activation levels, such as voting or activation spreading \cite{maes1989dynamics,payton1992whatever}. Behavior-based systems are related to the works of reactive robotics, particularly the {\it subsumption architecture} \cite{brooks2003robust}. Through embedding the robot’s controller with a collection of preprogrammed parallel condition–action rules or reflexes in a minimal internal state, it can achieve rapid real-time responses \cite{agre1987pengi,brooks1987asynchronous}.  

Specifically, task-achieving behaviors in the subsumption architecture are represented as separate layers. Individual layers work on individual goals concurrently and asynchronously. At the lowest level, each behavior is represented by a behavioral module. Where each behavioral module performs an action and is responsible for its own perception of the world, and there is no global memory, bus, or clock. (Fig. \ref{fig:bm}) that encapsulates a specific behavioral transformation function. Stimulus or response signals can be suppressed or inhibited by other active behaviors. A reset input is also used to restore the behavior to its initial state. A priority hierarchy fixes the topology of behavioral layers. The lower levels in the architecture are unaware of the higher levels, which provides the basis for incremental design. Higher-level competencies are added on top of an already functioning control system without modifying the lower levels. Fig. \ref{fig:bbr_layers} shows a simple mobile robot with three behavioral layers. 

To summarize, the key aspects for the design of subsumption-style robots in BRR are {\it situatedness} and {\it embodiment}. {\it Situatedness} refers to the robot’s ability to sense its current surroundings and avoid using abstract representations, while {\it embodiment} insists that robots be physical creatures and thus experience the world directly rather than through simulation \cite{brooks1999cambrian,arkin1998behavior}. To some extent, BBR promotes the integration of robotics and cognitive science, and is a traditional embedded AI system \cite{yifan2025embodied} considered an example of weak artificial intelligence.
\begin{table*}
\caption{Competency Areas and System Types of Typical Cognitive Architectures}
\begin{center}
\begin{threeparttable} 
\scriptsize
\renewcommand{\arraystretch}{1.5}
\begin{tabular}{cccccccccc} 
\hline 
\textbf{Cognitive Architectures} & \textbf{Perception} & \textbf{Attention} & \textbf{Motivation} & \textbf{Reasoning} & \textbf{Learning} & \textbf{Actuation} & \textbf{Interaction} & \textbf{Creativity} & \textbf{Type}\tnote{*}\\
\hline
\makecell[l]{Soar: \cite{laird2019soar} \\ [3pt] ATC-R: \cite{anderson2001tower} \\ [3pt] CLARION: \cite{sun2005interaction} \\ [3pt] NARS: \cite{wang2013natural} \\ [3pt] LIDA: \cite{faghihi2012lida} \\ [3pt] SiMA: \cite{schaat2015interdisciplinary} \\ [3pt] Sigma: \cite{pynadath2014reinforcement} \\ [3pt] CogPrime: \cite{goertzel2014cognitive} \\ [3pt] ARCADIA: \cite{bridewell2015incremental} \\ [3pt] STAR: \cite{tsotsos2016attention} \\ [3pt] CELTS: \cite{faghihi2011emotional} \\ [3pt] PRODIGY: \cite{epstein2004metaknowledge} \\ [3pt] MusiCog: \cite{maxwell2012musicog} \\ [3pt] ATLANTIS: \cite{gat1992integrating} \\ [3pt] CoSy: \cite{pacchierotti2006embodied} \\ [3pt] 3T: \cite{firby1995architecture} \\ [3pt] Leabra: \cite{o1996leabra} \\ [3pt] SASE: \cite{weng2002developmental}}&
\makecell[c]{$\bigstar$ \\ [3pt] $\bigstar$ \\ [3pt] $\bigstar$ \\ [3pt] $\bigstar$ \\ [3pt] $\bigstar$ \\ [3pt] $\bigstar$  \\ [3pt] $\bigstar$  \\ [3pt] $\bigstar$  \\ [3pt] $\bigstar$ \\ [3pt] $\bigstar$ \\ [3pt] $\bigstar$ \\ [3pt] $\bigstar$  \\ [3pt] $\bigstar$  \\ [3pt] $\bigstar$  \\ [3pt] $\bigstar$ \\ [3pt] $\bigstar$ \\ [3pt] $\bigstar$ \\ [3pt] $\bigstar$}&
\makecell[c]{$\bigstar$ \\ [3pt] $\bigstar$ \\ [3pt] $\bigstar$ \\ [3pt] $\bigstar$ \\ [3pt] $\bigstar$ \\ [3pt] $\bigstar$  \\ [3pt] $\bigstar$  \\ [3pt] $\bigstar$  \\ [3pt] $\bigstar$  \\ [3pt] $\bigstar$  \\ [3pt] $\bigstar$ \\ [3pt] $\bigstar$  \\ [3pt] $\bigstar$  \\ [3pt] $\bigstar$  \\ [3pt] $\bigstar$ \\ [3pt] $\bigstar$ \\ [3pt] $\bigstar$ \\ [3pt] $\bigstar$}&
\makecell[c]{$\bigstar$ \\ [3pt] $\bigstar$ \\ [3pt] $\bigstar$ \\ [3pt] $\bigstar$ \\ [3pt] - \\ [3pt] $\bigstar$  \\ [3pt] $\bigstar$  \\ [3pt] $\bigstar$  \\ [3pt] -  \\ [3pt]  - \\ [3pt] $\bigstar$ \\ [3pt] -  \\ [3pt] -  \\ [3pt] -  \\ [3pt] - \\ [3pt] - \\ [3pt] - \\ [3pt] -}&
\makecell[c]{$\bigstar$ \\ [3pt] $\bigstar$ \\ [3pt] $\bigstar$ \\ [3pt] $\bigstar$ \\ [3pt] $\bigstar$ \\ [3pt] $\bigstar$  \\ [3pt] $\bigstar$  \\ [3pt] $\bigstar$  \\ [3pt] $\bigstar$  \\ [3pt]  $\bigstar$ \\ [3pt] $\bigstar$ \\ [3pt] $\bigstar$  \\ [3pt] $\bigstar$  \\ [3pt] $\bigstar$  \\ [3pt] $\bigstar$ \\ [3pt] $\bigstar$ \\ [3pt] $\bigstar$ \\ [3pt] $\bigstar$}&
\makecell[c]{$\bigstar$ \\ [3pt] $\bigstar$ \\ [3pt] $\bigstar$ \\ [3pt] $\bigstar$ \\ [3pt] $\bigstar$ \\ [3pt] $\bigstar$  \\ [3pt] $\bigstar$  \\ [3pt] $\bigstar$  \\ [3pt]  - \\ [3pt]  - \\ [3pt] $\bigstar$ \\ [3pt] $\bigstar$  \\ [3pt] $\bigstar$  \\ [3pt] -  \\ [3pt] $\bigstar$ \\ [3pt] - \\ [3pt] $\bigstar$ \\ [3pt] $\bigstar$}&
\makecell[c]{$\bigstar$ \\ [3pt] $\bigstar$ \\ [3pt] $\bigstar$ \\ [3pt] - \\ [3pt] $\bigstar$ \\ [3pt] $\bigstar$  \\ [3pt] $\bigstar$  \\ [3pt] $\bigstar$  \\ [3pt] -  \\ [3pt] $\bigstar$  \\ [3pt] - \\ [3pt] $\bigstar$  \\ [3pt] -  \\ [3pt] $\bigstar$  \\ [3pt] $\bigstar$ \\ [3pt] $\bigstar$ \\ [3pt] $\bigstar$ \\ [3pt] $\bigstar$}&
\makecell[c]{$\bigstar$ \\ [3pt] $\bigstar$ \\ [3pt] $\bigstar$ \\ [3pt] - \\ [3pt] $\bigstar$ \\ [3pt] $\bigstar$  \\ [3pt] $\bigstar$  \\ [3pt] $\bigstar$  \\ [3pt] -  \\ [3pt] -  \\ [3pt] $\bigstar$ \\ [3pt] $\bigstar$  \\ [3pt] -  \\ [3pt] -  \\ [3pt] $\bigstar$ \\ [3pt] $\bigstar$ \\ [3pt] - \\ [3pt] $\bigstar$}&
\makecell[c]{ - \\ [3pt] - \\ [3pt] $\bigstar$ \\ [3pt] - \\ [3pt] - \\ [3pt] -  \\ [3pt]  -  \\ [3pt] -  \\ [3pt]  - \\ [3pt]  - \\ [3pt] - \\ [3pt] -  \\ [3pt] $\bigstar$  \\ [3pt] -  \\ [3pt] - \\ [3pt] - \\ [3pt] - \\ [3pt] -}&
\makecell[c]{ \textbf{HA} \\ [3pt] \textbf{HA} \\ [3pt] \textbf{HA} \\ [3pt] \textbf{HA} \\ [3pt] \textbf{HA} \\ [3pt] \textbf{HA}  \\ [3pt]  \textbf{HA}  \\ [3pt] \textbf{HA}  \\ [3pt]  \textbf{HA} \\ [3pt]  \textbf{HA} \\ [3pt] \textbf{HA} \\ [3pt] \textbf{SA}  \\ [3pt] \textbf{SA}  \\ [3pt] \textbf{HA}  \\ [3pt] \textbf{HA} \\ [3pt] \textbf{HA} \\ [3pt] \textbf{EA} \\ [3pt] \textbf{EA}}
\\
\hline
\end{tabular}
\begin{tablenotes}
\item[*]\textbf{Note}: Symbolic Architectures: \textbf{SA}; Emergent Architectures: \textbf{EA}; Hybrid Architectures: \textbf{HA}.
\end{tablenotes}
\end{threeparttable} 
\end{center}
\label{table:ca_criteria}
\vspace{-5mm}
\end{table*}

\subsection{Cognitive Architectures}

Instead of studying individual low-level or simple behaviors to model robots’ interactions in BBS, cognitive architectures stand for more general and abstract aspects, such as perception, attention, motivation, reasoning, learning, and creativity, to embody AI agents as creatures. They are part of Artificial General Intelligence (AGI) research. The goal of the research in cognitive architectures is to model the human mind, eventually enabling us to build human-level artificial intelligence \cite{kotseruba202040}. Existing cognitive architectures have explored four perspectives of AI: mimicking human thinking styles, rational thinking, mimicking human behaviors, and rational actions \cite{russell1995modern}, which provide evidence for particular mechanisms succeeding in generating intelligent behaviors. A cognitive architecture reflects three aspects of a cognitive agent \cite{langley2009cognitive}, which are constant over time and across different application domains:
\begin{itemize}
    \item Having the short-term and long-term memories to store the agent’s beliefs, goals, experience, and knowledge;
    \item The components and their organization represent larger-scale mental structures in memories;
    \item The functional processes of corresponding structures involve actuators and learning mechanisms.
\end{itemize}

Specifically, a typical cognitive architecture usually contains several critical components, such as a perceptron for processing sensory inputs and translating them into meaningful information, an attention mechanism for selecting relevant stimuli, working memories for temporarily storing and manipulating information to perform tasks, long-term memories for storing and retrieving information through learning and experience, reasoning and decision making for processing information and making inferences to reach decisions, and action modules for executing motor actions that are necessary to achieve goals. Fig. \ref{fig:ca} illustrates a cognitive architecture with the autonomy and learning capacity to achieve lifelong open-ended learning autonomy (LOLA)\footnote{Lifelong open-ended learning autonomy emphasizes autonomous goal discovery, inherently tied to challenges like competence acquisition, curriculum learning, and adaptation in non-stationary environments \cite{sigaud2023definition,romero2025h}.}.
\begin{figure}
	\centering
    \includegraphics[width=\linewidth]{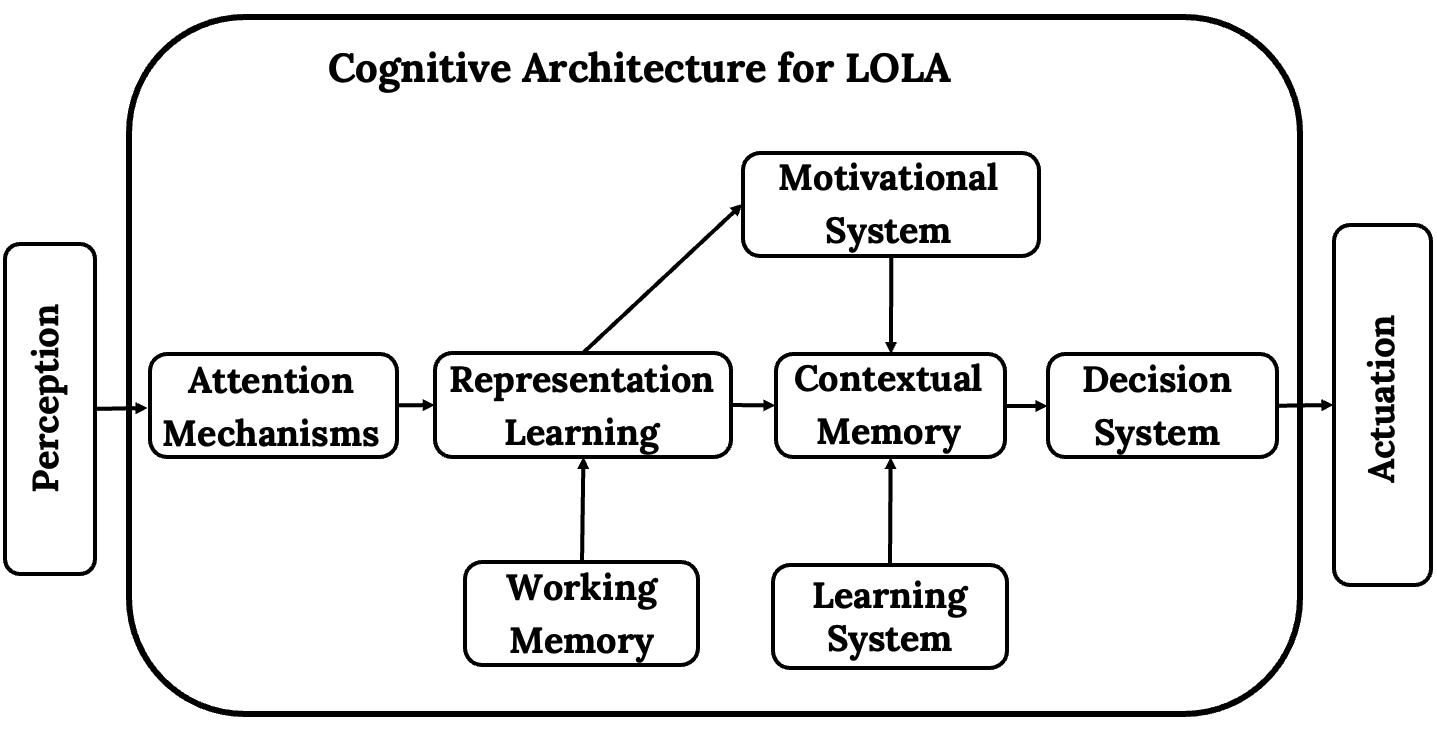}
    \caption{The illustration of a cognitive architecture reaching the autonomy and learning capacity necessary to achieve LOLA \cite{romero2023perspective}.}
    \label{fig:ca}
\end{figure}

\subsubsection{Criteria of Cognitive Architectures}
Due to the absence of a clear definition and a general theory of cognition, most cognitive architectures rely on particular premises and assumptions, making comparisons and evaluations of progress across architectures difficult \cite{kotseruba202040}. Some researchers have argued that current models of cognitive architecture do not adequately reflect the complexity of human cognition, thereby limiting machine intelligence's ability to replicate human decisions. Furthermore, they have not provided a theoretical framework for explaining how the mind's cognitive processes interrelate. Although the current cognitive architecture models can explain how individual cognitive processes work, they don't provide a comprehensive understanding of how these processes interact and shape human behavior.

To develop criteria for cognitive architecture, some researchers have done pioneering work. \cite{newell1980physical,newell1992precis} proposed the functional criteria, which include flexible behavior, real-time operation, rationality, a large knowledge base, learning, development, linguistic abilities, self-awareness, and brain realization. \cite{sun2004desiderata} extends the criteria to broader domains, including ecological, cognitive, bioevolutionary realism, adaptation, modularity, routineness, and synergistic interaction. Furthermore, some researchers proposed more practical solutions regarding it as a set of competencies and behaviors demonstrated by the system, such as perception, memory, attention, actuation, reasoning, learning, social interaction, planning, motivation, emotion, creativity, etc \cite{adams2012mapping}. Table. \ref{table:ca_criteria} presents typical cognitive architectures and their competency areas.

\subsubsection{Categories of Cognitive Architectures}
Based on previous literature reviews \cite{nakhaeinia2011review,kotseruba202040,gonzalez2025cognitive}, cognitive architectures can be regarded as control architectures and classified into three categories: {\it Symbolic}, {\it Emergent}, and {\it Hybrid}. Fig. \ref{fig:categories} illustrates the classification of cognitive architectures and their corresponding control architectures.

\paragraph{Symbolic Architectures:} Symbolic architectures are closely related to cognitivism, which views the human mind as processing information through symbolic representations and logical rules. They utilize formal languages to represent information and rules guiding decision-making. Their hierarchical structure can set high-level goals to organize task planning and execution. Like the deliberative architectures, symbolic architectures also focus on information processing based on symbolic knowledge. For example, robots can use symbolic information and logical rules to achieve high-level abstraction and long-term planning, such as route planning in unstructured environments, natural language processing, logical inference, and real-time decision-making \cite{laird1987soar}. Moreover, the properties of symbolic architectures make them particularly suitable for formal verification and safety-aware planning, thereby contributing to safety, verification, and explainability in robotic systems \cite{m2023model,bhuyan2024neuro}.

\paragraph{Emergent Architectures:} Unlike symbolic architectures based on symbolic representations or logical rules, emergent architectures build on complex behaviors and capabilities emerging from the interaction of multiple simple elements. On the other hand, reactive architectures are built on predefined sets of rules and behaviors specific to particular situations. Although they share features with reactive architectures in perception-based decision-making and adaptation, emergent architectures use neural networks and other machine learning techniques to generate behaviors, often learning and adapting based on experience and environmental feedback. Moreover, the enactive interpretation attribute causes them to adjust their behavior as they explore new situations or receive new data, which might sabotage the previously learned model. Recent studies unify views of perception, action, and learning based on the free-energy principle, serving as architectural principles for embodied agents \cite{friston2010free,pio2016active}.

\paragraph{Hybrid Architectures:} As we discussed above, symbolic architectures are good at well-defined problems, such as planning and reasoning. Their knowledge base is based on human interpretation and needs to be generated from initial states, which makes them hard to adapt to dynamic environments. Although emergent architectures are highly adaptable and exhibit enactive interpretation, they might degrade model performance by using sub-symbolic knowledge and exploring new data. To tackle those issues, hybrid architectures attempt to combine two frameworks into a single structure. They usually have a symbolic planner in charge of long-term planning, and several emergent modules that provide actions with greater adaptability. For example, the neuro-symbolic integration incorporates symbolic reasoning and planners with neural perception and learning modules, involving translating neural outputs to symbols and tightly coupled systems to interleave learning and reasoning \cite{bhuyan2024neuro}. Moreover, some foundation models, such as Large Language Models (LLMs) and multimodal Vision-Language Models (VLLs), accelerate the integration of perception, reasoning, and low-level skills by serving as high-level planners, communicators, and knowledge sources, generating action sequences or symbolic plans \cite{ahn2022can,huang2022inner,wang2025large,mon2025embodied,kannan2024smart}.
\begin{figure*}
	\centering
    \includegraphics[width=\linewidth]{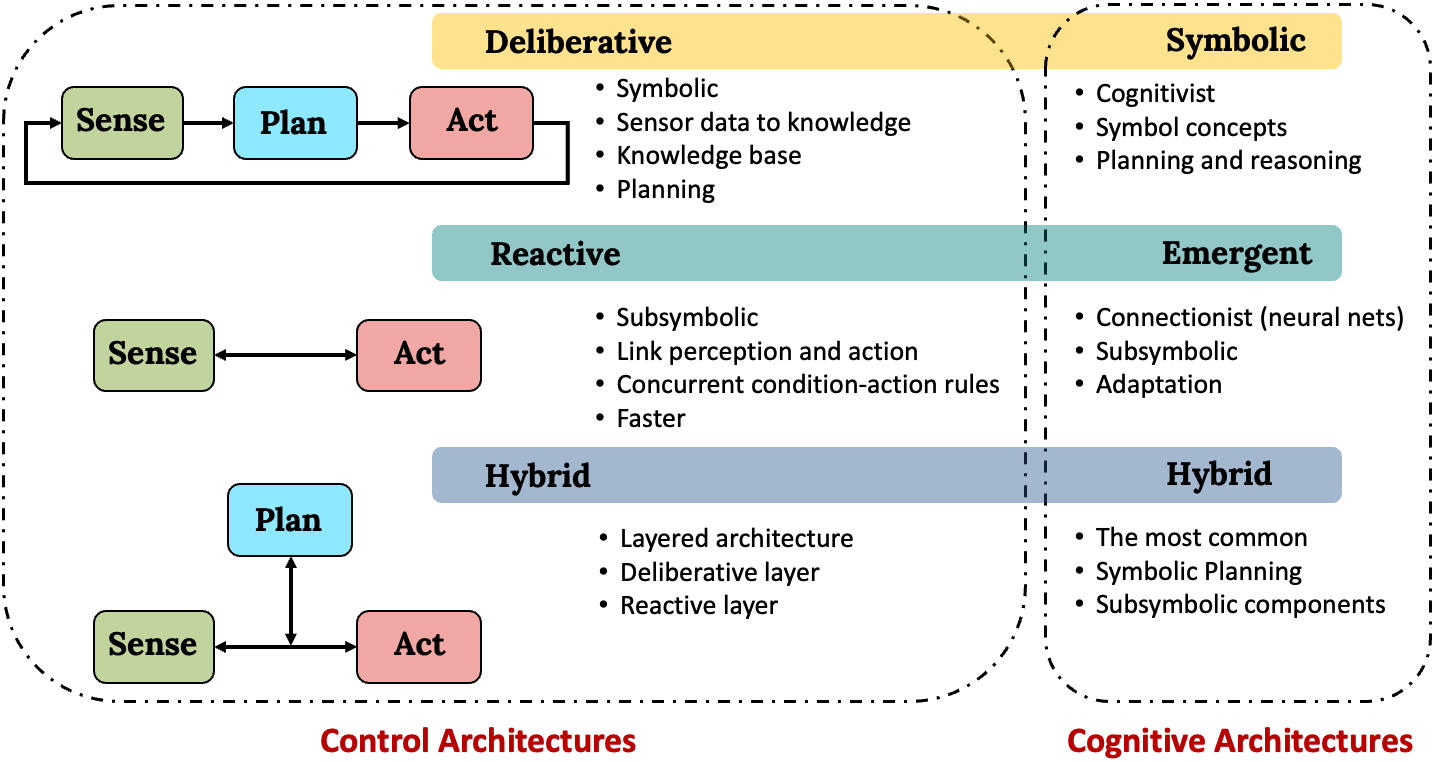}
    \caption{Illustration of the classification of cognitive architectures and their corresponding control architectures.}
    \label{fig:categories}
\end{figure*}

\subsubsection{Cognitive Robotics} Cognitive architectures play an essential role in the development of intelligent autonomous robots, including planning and control, decision-making, learning, and adaptation to dynamic environments, which lay the foundation for cognitive robotics. Specifically, developmental and cognitive robotics studies the mechanisms, architectures, and constraints that enable lifelong, open-ended learning of new skills and knowledge in embodied machines. They aim to establish cognitive functions through a synthetic approach to better understand higher cognitive behaviors in humans \cite{asada2009cognitive}, employing knowledge representation, reasoning, and learning techniques to support autonomous robots in dynamic and unstructured environments \cite{levesque2008cognitive}. In the early stage of cognitive robotics studies, various robots, such as mobile robots, robot arms, and humanoids, were used to investigate higher-order cognitive capabilities that mimic the functionalities of the human brain, including its internal structure, infrastructure, and social structures. 

\paragraph{Social Cognition: }

Social cognition refers to the ability to recognize and control the self in relation to others and to perceive and apply social signals in the interaction. It extends cognitive robotics into a more general and sophisticated domain that not only considers multi-robot exchange of information and knowledge and the acquisition of new skills from others, but also involves building reliable and sustainable relationships with humans in the long-term development. For example, research on socially interactive robots \cite{fong2003survey,breazeal2003toward} is interested in very high-level social interaction conducted by robotic agents, focusing on the body of research dedicated to designing task non-specific social cognition in robotic agents as part of fulfilling the goal of designing human-like cognition in robotic systems. Therefore, ``sociable robot" research on social cognition development in autonomous systems primarily focuses on mechanisms that enable robots to interact with humans in two main categories, including {\it Joint Attention} which refers to the ability to intentionally attend to an object/region of mutual interest
\cite{breazeal1998motivational,brooks1998cog,breazeal2002active,nagai2003constructive,hosoda2004acquisition}, and {\it Social Imitation} which includes sensori-motor skills for generating different types of social cues and task sequence learning during development \cite{breazeal2002robots,kuniyoshi2003visuo,shon2007towards,ito2004line,calinon2005goal,calinon2007learning}.

\paragraph{Research Directions: }
The core challenge of cognitive robotics is the gap between the robot’s knowledge and its reasoning capabilities \cite{toberg2024commonsense}, which makes automatic acquisition and deployment more difficult. Especially, let robots share common-sense knowledge (CSK) with humans in everyday tasks, such as knowledge about human desires, physics, causality, emotion, properties, and relationships \cite{gupta2004common}. This knowledge relates to the concept of {\it Intuitive Psychology} \cite{lake2017building}, which holds that an agent understands other agents with a mental state similar to their own, and can express and interpret this to achieve mutual understanding of their intentions and goals. One possible solution is {\it Physical Embodiment}, which means that the agent’s physical body specifies the constraints on the interaction between the agent and its environment that generate the rich contents of its process or consequences \cite{brooks1991intelligence,asada1999cooperative,asada2001cognitive,vernon2007survey,kuniyoshi2007emergence,asada2009cognitive}. Inspired by the concept, researchers build artificial systems capable of acquiring motor and cognitive capabilities through interaction with the environment, in line with human development \cite{asada2009cognitive}. Another approach gradually implemented in cognitive robotics is {\it predictive processing} (PP), a top-down approach that aims to integrate perception, cognition, and action as a single inference process \cite{ciria2021predictive}, grounded in the free-energy principle \cite{buckley2017free} and frameworks such as predictive coding, active inference, and perceptual inference. 

Furthermore, the development of machine learning techniques has led to a range of learning paradigms, including imitation learning, transfer learning, representation learning, reinforcement learning, and curriculum learning \cite{tani2003learning}. It provides systematic approaches, ranging from mastering lower-level basic skills, such as trajectory planning and multi-DOF (Degrees of Freedom) control \cite{yang2024bayesian}, to more sophisticated systems, including modules for visual attention, speech recognition, and the integration of visual and linguistic inputs for instructing robots to grasp everyday objects \cite{steil2006recent}. As various techniques, such as AI, quantum computing, wireless sensors, and cloud computing, have rapidly developed, trends in cognitive robotics are presented below \cite{tawiah2022machine}:
\begin{itemize}
    \item Synthesizing computational models for high-order cognitive skills through neuroscience and behavioural psychology in AI agents;
    \item Utilizing motivations or goal-directed mechanisms to balance exploration and exploitation in task spaces;
    \item Using predictive coding mechanisms to synthesise higher-order cognitive behaviours;
    \item Social behaviours studies in robotic swarms;
    \item Large implementation of networked and cloud robotics and cyber-physical systems in higher-order robotic research, like social robots.
\end{itemize}

\subsection{Utility Theory}

\subsubsection{Basic Concepts}
The dominant approach to modeling an agent's interests or needs is {\it utility theory}. This theoretical approach aims to quantify an agent's degree of preference across a set of available alternatives and understand how these preferences change when an agent faces uncertainty about which alternative he will receive \cite{shoham2008multiagent}. 
      
To describe the interactions between multiple utility-theoretic agents, we use the specific {\it utility function} to analyze their preference and rational action. The utility function is a mapping from states of the world to real numbers, which are interpreted as measures of an agent's level of happiness (needs) in the given states. If the agent is uncertain about its current state, the utility is defined as the {\it expected value} of its utility function for the appropriate probability distribution over states \cite{shoham2008multiagent}. From the perspective of the connection between a decision maker and its preference, a decision maker would rather implement a more preferred alternative (act, course of action, strategy) than one that is less preferred \cite{fishburn1970utility}.

Furthermore, {\it expected utility theory} states that the decision maker choose between risky or uncertain prospects by comparing their expected utility values, that is, the weighted sums obtained by adding the utility values of outcomes multiplied by their respective probabilities \cite{mongin1998expected}. There are two kinds of theories based on the standard distinction between risk and uncertainty: {\it Subjective expected utility theory} (SEUT) and {\it Von Neumann–Morgenstern theory} (VNMT). 
\begin{figure*}
	\centering
    \includegraphics[width=\linewidth]{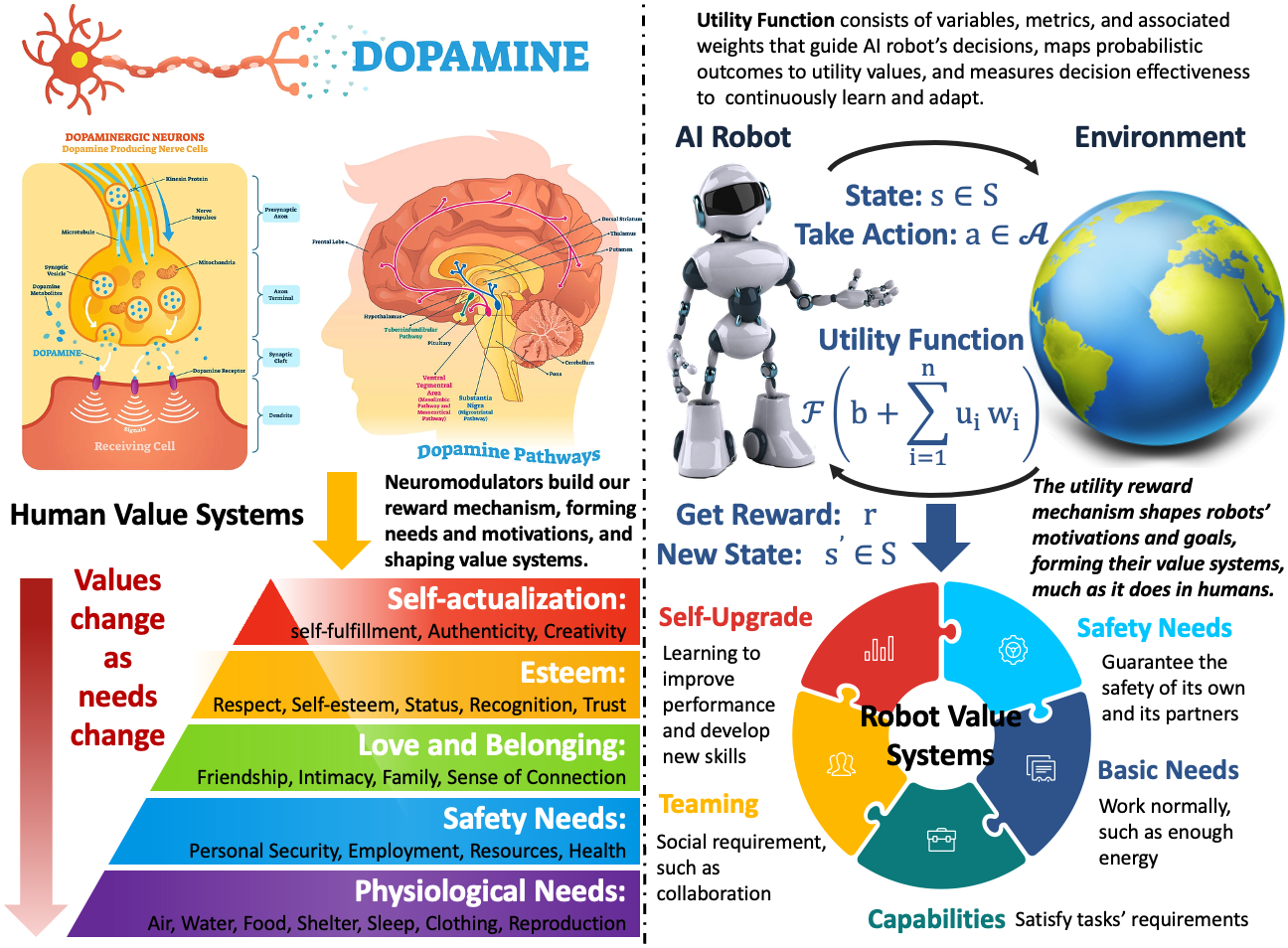}
    \caption{Illustration of the value systems of human based on neuromodulators \cite{guy2024dopamine} and the AI robot's value systems from utility theory perspective.}
    \label{fig:value_systems}
\end{figure*}

\paragraph{Subjective expected utility theory:}
The theory of subjective expected utility describes the attractiveness of an economic opportunity as perceived by a decision-maker in the presence of risk, which combines two subjective concepts: a personal utility function and a personal probability distribution (usually based on Bayesian probability utility theory) \cite{savage1972foundations,Karni2017}.

In SEUT, if the decision maker adheres to axioms of rationality, believing an uncertain event has possible outcomes $\{ x_i \}$ each with utility $u(x_i)$, then the person's choices can be explained as arising from this utility function combined with the subjective belief that there is a probability of each outcome, $P(x_{i})$. And the subjective expected utility is Eq. \eqref{sub_utility1}.
\begin{equation}
\begin{split}
    \mathbb{E} \left[ u(X) \right] = \sum_i u(x_i) \cdot P(x_i)
\label{sub_utility1}
\end{split}
\end{equation}

The decision relies on the higher subjective expected utility, which means agents may make different decisions because of their various utility functions or beliefs about the probabilities of different outcomes. Moreover, \cite{mongin1998utility} further discussed the connection between ethical and utility theory describing the complex social behaviors in human.

\paragraph{Von Neumann–Morgenstern utility theory:}
In decision theory, the {\it Von Neumann–Morgenstern utility theory} (VNMT) is also known as the {\it expected utility hypothesis}, which shows that, under certain axioms\footnote{The four axioms of VNM-rationality are {\it completeness}, {\it transitivity}, {\it continuity}, and {\it independence}. \cite{von2007theory}} of rational behavior, a decision-maker faced with risky (probabilistic) outcomes of different choices will behave as if the agent is maximizing the expected value of some function defined over the potential outcomes at some specified point in the future \cite{von2007theory}.

\paragraph{Maslow's hierarchy of human needs:} From the psychological perspective, {\it Maslow's hierarchy of needs} \cite{maslow1943theory} is used to study how humans intrinsically partake in behavioral motivation. It is often portrayed in the shape of a pyramid, with the largest, most fundamental needs at the bottom, and the need for self-actualization and transcendence at the top. In other words, the idea is that individuals' most basic needs must be met before they become motivated to achieve higher-level needs. Maslow used the terms {\it physiological}, {\it safety}, {\it belonging and love}, {\it social needs or esteem}, and {\it self-actualization} to describe the pattern through which human motivations generally move and the conditional dependent relationships between each level's needs. In other words, if the agent's motivation wants to arise at the next stage, the current level's {\it needs (utilities)} must satisfy a certain amount. Furthermore, for more advanced intelligent agents, such as human beings, many motivations from different levels of needs can occur simultaneously. Instead of focusing on a certain need at any time, \cite{maslow1981motivation} stated that a specific need might dominate the decision at a certain time, and the likelihood of the dominant need will follow the order presented in the needs hierarchy.

Generally speaking, if we can build a model describing the agent's hierarchy of needs based on the {\it expected utility theory}, we can develop a more intelligent single agent and MAS representing more complex behaviors to adapt to various dynamic and complex environments.

\subsubsection{Experienced Utility in Cognitive Process}

People make decisions to gain pleasure and avoid pain \cite{bentham1996collected}. In the view of neoclassical economics, people try to maximize utility to fit their individual preferences or needs, which starts with a hedonic approach to utility that aims to produce the most pleasure and the least pain \cite{berridge2014experienced1}. This view can also be explained in terms of neuroscience. Neuromodulators in the central nervous system include serotonin, acetylcholine, dopamine, and norepinephrine, which transmit signals across multiple neurons. They build our reward mechanism (Fig. \ref{fig:value_systems}) to induce subjective feelings of pleasure and contribute to positive emotions \cite{schultz2000multiple}. For an individual, a good decision is to maximize the hedonic or pleasurable experience of the chosen outcome
\footnote{Outcomes that generate a pleasure impact elicit a constellation of objective responses (including affective behavioral reactions, physiological autonomic, and brain limbic reactions) as well as in humans at least, subjective feelings reported as pleasure.}, termed {\it Experienced Utility} \cite{fischhoff2021experienced}. Experienced utility or subjective pleasure involves cognitive construal of future goals \cite{kahneman1997back}. The process generates our motivation, needs, and beliefs \cite{eccles2002motivational}, described as a cognitive form of anticipated utility. From a neuroeconomic perspective, decision-making based on experienced utility is associated with widespread neural activation across a diffuse circuit network involving many brain structures, which helps us better understand the relationships between cognition and the brain \cite{glimcher2013neuroeconomics}.

\subsubsection{Utility Value Systems of Cognition}

A value system facilitates the capacity of a biological brain to increase the likelihood of neural responses to an external phenomenon \cite{begum2009computational}. Value presents a subjective point of view to measure the effort of an agent willing to expend to obtain rewards or to avoid punishment from goals. Due to its biological inspiration, it is termed innate, inherent, or motivational value \cite{merrick2017value}, which reflects a subjective evaluation of the sensory space. In the life of an agent, the agent starts from inherent values, and its acquired values are modified through interaction with various environments and knowledge accumulation in learning. So the acquired value is thus activity-dependent and allows the value system to become sensitive to stimuli that are not able to trigger a value-related response by themselves \cite{sporns2010value}. \cite{sporns2000plasticity} explains that a vertebrate’s value system is realised in part through the roles of neuromodulators and a form of brain plasticity called long-term potentiation, which develops acquired value systems based on the reward mechanism formed through neuromodulators, as we discussed above (Fig. \ref{fig:value_systems}).
To evaluate an agent's innate values and motivations, utility theory aims to quantify an agent's degree of preference across a set of available alternatives and understand how these preferences change when the agent faces uncertainty about which alternative it will receive \cite{shoham2008multiagent,glimcher2013neuroeconomics}. Furthermore, modern utility theory has its origins in the theory of expected value, and the value of any course of action can be determined by multiplying the gain from that action by its likelihood \cite{glimcher2005physiological}.

Specifically, neurons process utility rewards derived from environmental information via distinct brain structures. They detect the expectation of imminent rewards or anticipated utility in the value and identity of encountered stimuli, as reflected in the control of goal-directed behaviour. The stimuli will be labelled with a positive value proportional to the degree of the reward signal. This information helps construct neuronal representations that enable subjects to anticipate future utility rewards based on prior experience. Through cognitive appraisals of the experience utility, the individual can perform subjective cost-benefit calculations to determine the level of effort required and execute effortful behavioural responses. To balance the contingencies and utility rewards, it naturally adopts adaptive behaviour to respond to the changing environments. This process will gradually form its unique value direction and preferences, represented as a distribution of needs \cite{yang2025innate}, termed value systems \cite{schultz2000multiple}. 

To summary, utility serves as a unified concept in value systems to describe the capacity of action, behavior, and strategy to satisfy the agents' desires or needs in their decision-making. 


\subsection{Utility Theory based Cognitive Modeling in Robotics}

\subsubsection{Value Systems of Cognitive Robots}
\begin{figure}
	\centering
    \includegraphics[width=\linewidth]{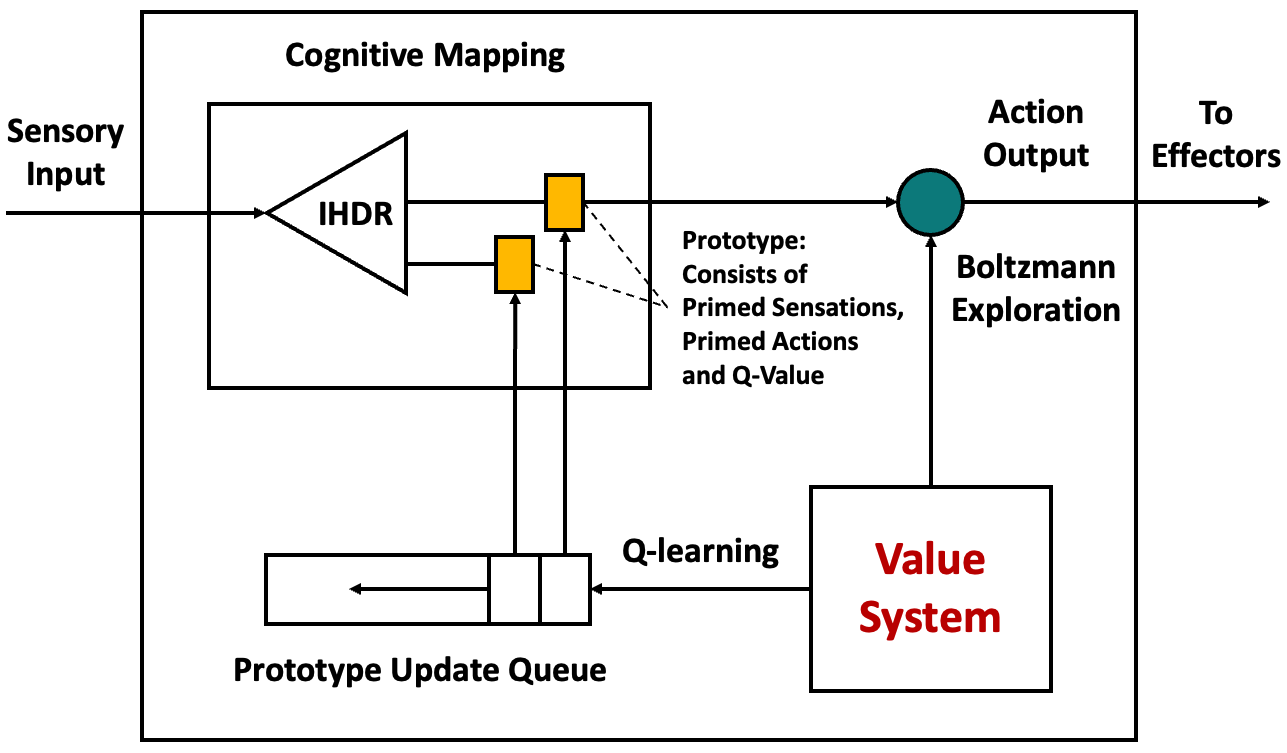}
    \caption{The illustration of the value system in the cognitive architecture of SAIL robot \cite{huang2002novelty}.}
    \label{fig:sail}
\end{figure}

\begin{table*}
\caption{Aspects of the Value System Design for Cognitive Robots}
\begin{center}
\begin{threeparttable} 
\scriptsize
\renewcommand{\arraystretch}{1.5}
\begin{tabular}{cccccc} 
\hline 
\textbf{References} & \textbf{Robot} & \textbf{Value Type}\tnote{1} & \textbf{Data Structure}\tnote{2} & \textbf{Initial Strategy}\tnote{3} & \textbf{Update Strategy}\tnote{4} \\
\hline
\cite{sporns2000plasticity} & NOMAD & \makecell[c]{Extrinsic: \\ [2pt] Taste} & Neural Network & Experience & Synaptic Rules \\
\hline
\cite{prescott2006robot} & Kephera & \makecell[c]{Extrinsic: \\ [2pt] Fear \& Hunger} & Feedforward Networks & Predefined & Euler method \\
\hline
\cite{cox2009neuromodulation} & CARL-1 & \makecell[c]{Intrinsic: \\ [2pt] Explore} & Simulated Neurons & Random & Weight Rules\\
\hline
\cite{bolado2013biologically} & Simulation & \makecell[c]{Intrinsic: \\ [2pt] Novelty} & Neural Units & Predefined & Multiple \\
\hline
\cite{merrick2010modeling} & \makecell[c]{Lego Mindstorms \\ [2pt] NXT} & \makecell[c]{Extrinsic: \\ [2pt] Behavior Cycles} & Neural
Network & Random & Q-Learning \\
\hline
\cite{martius2013information}& Simulation & \makecell[c]{Extrinsic: \\ [2pt] Empowerment} & Neural Network & Random & \makecell[c]{Information \\ [2pt] Maximisation} \\
\hline
\cite{huang2002novelty} & SAIL & \makecell[c]{Intrinsic: \\ [2pt] Novelty} & Regression Tree & Random & Q-Learning \\
\hline
\cite{doya2005cyber}& Cyber Rodent & \makecell[c]{Extrinsic: \\ [2pt] Reproduction} & Various & Various & \makecell[c]{Reinforcement \\ [2pt] Learning (RL)}\\
\hline
\cite{frank2014curiosity} & iCub Humanoid & \makecell[c]{Intrinsic: \\ [2pt] Curiosity} & \makecell[c]{Markov decision \\ [3pt] process (MDP)} & \makecell[c]{MoBeE \\ [2pt] \cite{frank2012modular}} & RL \\
\hline
\cite{lee2007staged} & Robot Arm & \makecell[c]{Intrinsic: \\ [2pt] Novelty} & Arrays & --- & Mapping \\
\hline
\cite{ku2015error} & uBot-6 & \makecell[c]{Intrinsic: \\ [2pt] Surprise} & \makecell[c]{Aspect Transition Graph \\ (ATG)} & \makecell[c]{Initial ATG \\ [2pt] Provided} & \makecell[c]{Recursive Bayesian \\ [2pt] Estimation}\\
\hline
\cite{santucci2016grail} & Simulation & \makecell[c]{Extrinsic: \\ [2pt] Changing Environment} & Multiple & Multiple & Multiple \\
\hline
\cite{fiore2014keep} & Simulation & \makecell[c]{Extrinsic: \\ [2pt] Changing Light} & Leaky Integrator Units & Experience & Hebbian Learning \\
\hline
\cite{gordon2012curious} & Lego Mindstorms Arm & \makecell[c]{Extrinsic: \\ [2pt] Predicting Error} & Neural Network & Random & iNAC\tnote{5} \\
\hline
\cite{oudeyer2007intrinsic} & Sony AIBO dog & \makecell[c]{Intrinsic: \\ [2pt] Competence Progress} & Exemplars & Random & \makecell[c]{Nearest Neighbour \\ [2pt] Algorithm} \\
\hline
\cite{ngo2012learning} & Katana Robot Arm & \makecell[c]{Intrinsic: \\ [2pt] Learning Progress} & Parametric Representation & Predefined & LSPI\tnote{6} \\
\hline
\cite{craye2015exploration} & Kinect & \makecell[c]{Intrinsic: \\ [2pt] Novelty} & Random forest & Random & Decision Tree \\
\hline
\makecell[c]{\cite{yang2019self} \\ [2pt] \cite{yang2020hierarchical}}& Simulation & \makecell[c]{Combined: \\ [2pt] Energy \& Saftey \& Tasks} & Needs Hierarchy & Predefined & Behavior trees \\
\hline
\cite{yang2021can} & Simulation & \makecell[c]{Combined: \\ [2pt] Energy \& Victim} & Needs Hierarchy & Predefined & Relative Entropy \\
\hline
\cite{yang2022game} & \makecell[c]{Robotarium \\ [2pt] \cite{pickem2017robotarium}} & \makecell[c]{Combined: \\ [2pt] Survival \& Pursuit} & Needs Hierarchy & Predefined & GUT\tnote{7} \\
\hline
\cite{yang2025innate} & Simulation & \makecell[c]{Combined: \\ [2pt] Survival \& Tasks} & \makecell[c]{Needs Hierarchy \& \\ [2pt] Neural Network} & Predefined & \makecell[c]{Innate-Values-Driven \\ [2pt] RL (IVRL)} \\
\hline
\end{tabular}
\begin{tablenotes}
\setlength{\parskip}{1ex}
\item[1] \textbf{Value Type}: Intrinsic Motivations, Extrinsic Motivations, and Combined Motivations.
\item[2] \textbf{Data Structure}: The type of value network is represented in a value system.
\item[3] \textbf{Initial Strategy}: The initial selected values affect early behaviors and innate values.
\item[4] \textbf{Update Strategy}: The mathematical process updates the acquired values.
\item[5] \textbf{iNAC}: Incremental atural actor-critic \cite{bhatnagar2007incremental}
\item[6] \textbf{LSPI}: Least squares policy iteration \cite{lagoudakis2003least}
\item[7] \textbf{GUT}: Game-theoretic utility tree \cite{yang2023hierarchical} 
\end{tablenotes}
\end{threeparttable} 
\end{center}
\label{table:value_design}
\vspace{-5mm}
\end{table*}

The value measures an agent's effort to obtain a reward or avoid punishment. It is not hard-wired for an AI agent (like robots), even a biological entity, and the specific value system achieved through experience reflects an agent's subjective evaluation of the sensory space. And the value mechanisms usually have been defined as the {\it expected values}, particularly in uncertain environments. Moreover, the innate value reflects an agent's subjective evaluation of the sensory space, but the acquired value is shaped through experience during its development. 
A value system plays a crucial role in developing human-like intelligence in robotic systems, reflecting a robot's self-awareness, trust, ethics, knowledge, motivations, and needs.
\cite{begum2009computational} defines the artificial value system for the autonomous robot as a mechanism that can reflect the effects of robots’ experience on future behaviors, such as perception, planning, exploration, and learning. Fig. \ref{fig:sail} illustrates that \cite{huang2002novelty} introduced the modulating effect of the value system to guide the behavior of an autonomous system in the SAIL robot cognitive architecture. Moreover, the robotic value system demonstrates its capacity to plan action upon detecting salient stimuli, performing action planning after analyzing its internal and external context \cite{huang2004value,sporns2002neuromodulation}. Here, the robotic internal context describes its knowledge developed through past experiences, and the external context expresses its preference for a planned action in the current environment.

Existing computational value systems exhibit various distinctions between value and reward, which we can abstract as a network of nodes and edges in the fig. \ref{fig:value_arch}.
Specifically, the input nodes receive information from the external environment, and the output nodes will trigger a behavioural response by the robot. The edge between each pair of nodes describes the strength of the connection, represented by a scalar weight. The path between the input and output nodes corresponds to the total strength through the value appraisal, which indicates the likelihood of a certain behavioural reaction to an external stimulus detected by input nodes. Therefore, a value system can be regarded as a sparse or dense network, presented as generic graph-like structures, such as a tree or tabular structure. Moreover, the Internal nodes can be neurons, decision nodes, observations, contexts, clusters, or parameters. Through updating different formulae, the system can adjust the reinforcement or decay of weighted edge connections.

Many computational models involve the term ``value'' \cite{friston2012value}, such as innate values in the agent's initial state and acquired values throughout the agent's life \cite{sporns2009value}. Considering primates' value systems as a benchmark, \cite{schultz2000multiple,edelman1978group} summarize the properties of an artificial value system for cognitive robots as follows:
\begin{figure}
	\centering
    \includegraphics[width=\linewidth]{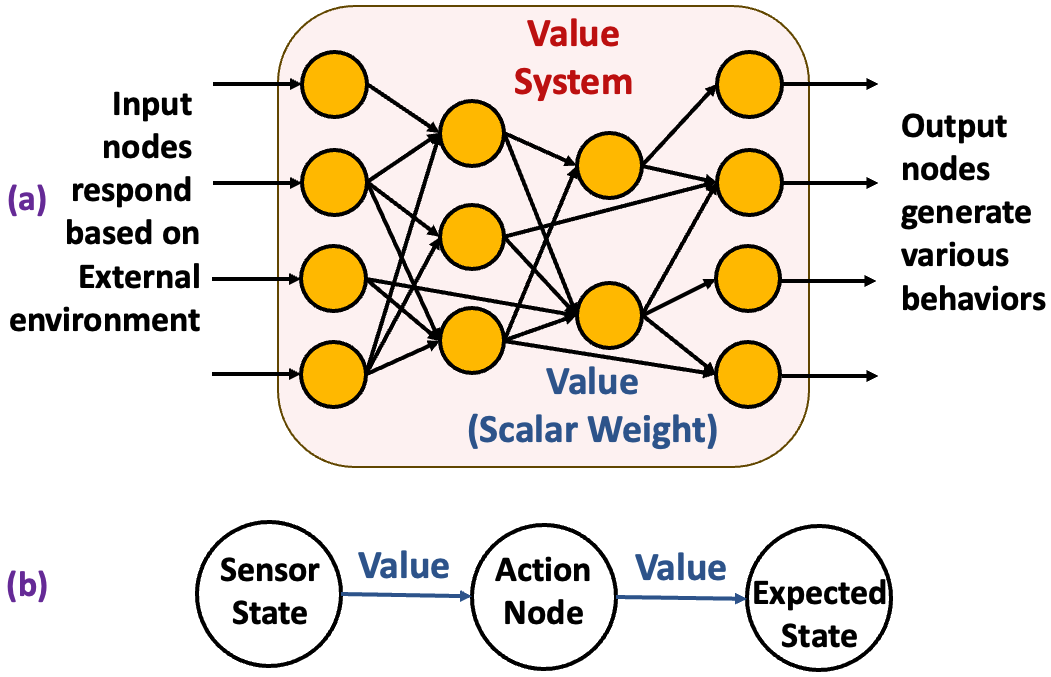}
    \caption{The illustration of the value system as a network of nodes and edges \cite{merrick2017value}.}
    \label{fig:value_arch}
\end{figure}

\begin{itemize}
    \item Prediction: According to external information, the value system can apply reasoning and knowledge to analyze the consequences of an action before executing it. It means that the value network (Fig. \ref{fig:value_arch}(a)) should have paths to link sensory stimuli, behavioral responses, and expected subsequent sensory stimuli (Fig. \ref{fig:value_arch}(b)).
    \item Task non-specific: An artificial value system should guide a robot to learn in a general and task-agnostic manner without human intervention. In the early stages, robots can receive limited support in developing their value systems. However, the systems need to have the capability to self-evolve in response to interactive environments, much as humans do\footnote{Instructional scaffolding is a widely used mechanism in primates (especially humans) to boost value system development during the very early stage, but the supervision provided through scaffolding is expected to decay gradually to facilitate natural cognitive development \cite{vygotsky2012thought}.}. 
    \item Self-regulation and Development: Self-regulation explains how people consciously manage their thoughts, emotions, and behaviors to achieve personal goals, which involves emotional regulation, behavioral control, and cognitive management \cite{karoly1993mechanisms}. It helps us develop reasonable value systems that adapt to current environments. Like a biological entity, the artificial agent also needs a self-regulation mechanism, including both innate and acquired components, to develop suitable value systems for sustainable development. Innate value is usually modeled as a prior over utility values, assigning different weights to actions. Correspondingly, acquired value will be generated over the long term throughout the interactive process. An artificial value system must have the plasticity to continuously encode feedback from its interactions with the environment, and the experience of this encoding should facilitate the emergence of adaptive behavior.
    \item Value-based learning: Learning plays a critical role throughout the development of artificial value systems, which should be able to generate modulating signals to adapt various learning parameters, such as bias (positive or negative) and learning rate \cite{begum2009computational}.
\end{itemize}

Table. \ref{table:value_design} lists design aspects and summarizes the corresponding research on value system design for autonomous robots. However, studies on designing artificial value systems with the goal of developing human-like intelligence in robotic systems remain very limited, especially in modeling human-like motivation mechanism in robots, quantifying value feedback from environments related to robotic current status and further development, and building self-upgrade artificial value systems in robots to adapt to dynamic changing environments. In the following parts, we provide the current methology about motivation modeling in robots from reviewed papers.

\subsubsection{Basic Concepts in Agent's Motivation}

From the organisms’ perspective, \cite{panksepp2004affective} defines the motivations of an intelligent agent as having three important functions: {\it Selection} drives the system to choose a suitable behavior to achieve the most important current needs/goals; {\it Energy} provides enough power to execute the selected behavior; and {\it Learning} generates learning signals to change behavior. Furthermore, \cite{baldassarre2011intrinsic,baldassarre2013intrinsically} proposed a more operational definition: regarding functions, extrinsic motivations have the overall function of driving behavior and learning for the acquisition of material resources, like biological needs: hunger and thirst. In contrast, intrinsic motivations are processes that can drive the acquisition of knowledge and skills in the absence of extrinsic motivations, such as curiosity, surprise, novelty, and success at accomplishing the agent’s own goals. In cognitive robotics, researchers define intrinsic motivations as a means of reward to influence the accumulation of value. Different approaches include using only intrinsic rewards or combining intrinsic and extrinsic rewards in various studies.    

However, the traditional AI agent lacks specific goal-setting to drive its motivations. Therefore, an intelligent cognitive architecture needs to involve the motivational system, integrating a series of mechanisms to support the agent's self-discovery and self-selection of goals based on the drive vector established by evolution or by the system designer.
Supposing AI agents can sense various environments, monitor their behaviors, and get internal and external information through perception $P(t)$ in the given time. And the data can generally be categorized into two classes: different variables in the environments (like light, sound, color, etc.) and agents' physical structure and operation (such as arm extension, contact in leg, etc), the operation of the cognitive processes within the robot (learning error, perceptual novelty, etc.) \cite{romero2019simplifying}. Then, the set of values transmits to the agent's actuators making up the actuation vector $A(t)$ in a given time. Furthermore, we can define these terms to help frame and understand related concepts in the cognitive modeling of motivation for the AI agent.

\begin{itemize}
    \item {\it Goal}: An agent obtains utilities in a perceptual state $P(t)$.
    \item {\it Innate Value/needs/Motivation/Drive}: An agent has the plasticity to encode the feedback received from its interactive environments continuously, and the encoding of experience should facilitate the emergence of adaptive behaviors \cite{merrick2017value}.
    \item {\it Strategy}: A strategy describes the general plan of an agent achieving short-term or long-term goals under uncertainty, which involves setting sub-goals and priorities, determining action sequences to fulfill the tasks, and mobilizing resources to execute the actions \cite{freedman2015strategy}.
    \item {\it Utility}: The benefit of using a strategy is to achieve a goal satisfying a specific innate value and needs. The utility is determined by the system's interaction with the environment and is not known a priori, traditionally called {\it Reward} in the reinforcement learning field \cite{romero2019simplifying}.
    \item {\it Expected Utility} ($e_u$): The probability of obtaining utility starts from a given perceptual state $P(t)$, which is modulated by the amount of utility obtained.
    \item {\it Utility Model}: A function provides the expected utility for any point in state space, which establishes the relationship between the real utility obtained from the interaction and how to achieve the goal.
    \item {\it Value Function (VF)}: A specific utility model expresses the expected utility for any perceptual state $P(t)$:
    \begin{equation}
        \begin{split}
            e_u(t+1) = VF\left(P(t+1) \right)
        \label{value_function}
        \end{split}
    \end{equation}
    \item {\it Episode}: It describes a circle of the interaction between agents and environments:
     \begin{equation}
        \begin{split}
            episode = \left\{P(t); A(t); P(t+1; e_u(t+1) \right\}
        \label{episode}
        \end{split}
    \end{equation}   
\end{itemize}
\begin{figure}
    \centering
    \includegraphics[width=0.9\linewidth]{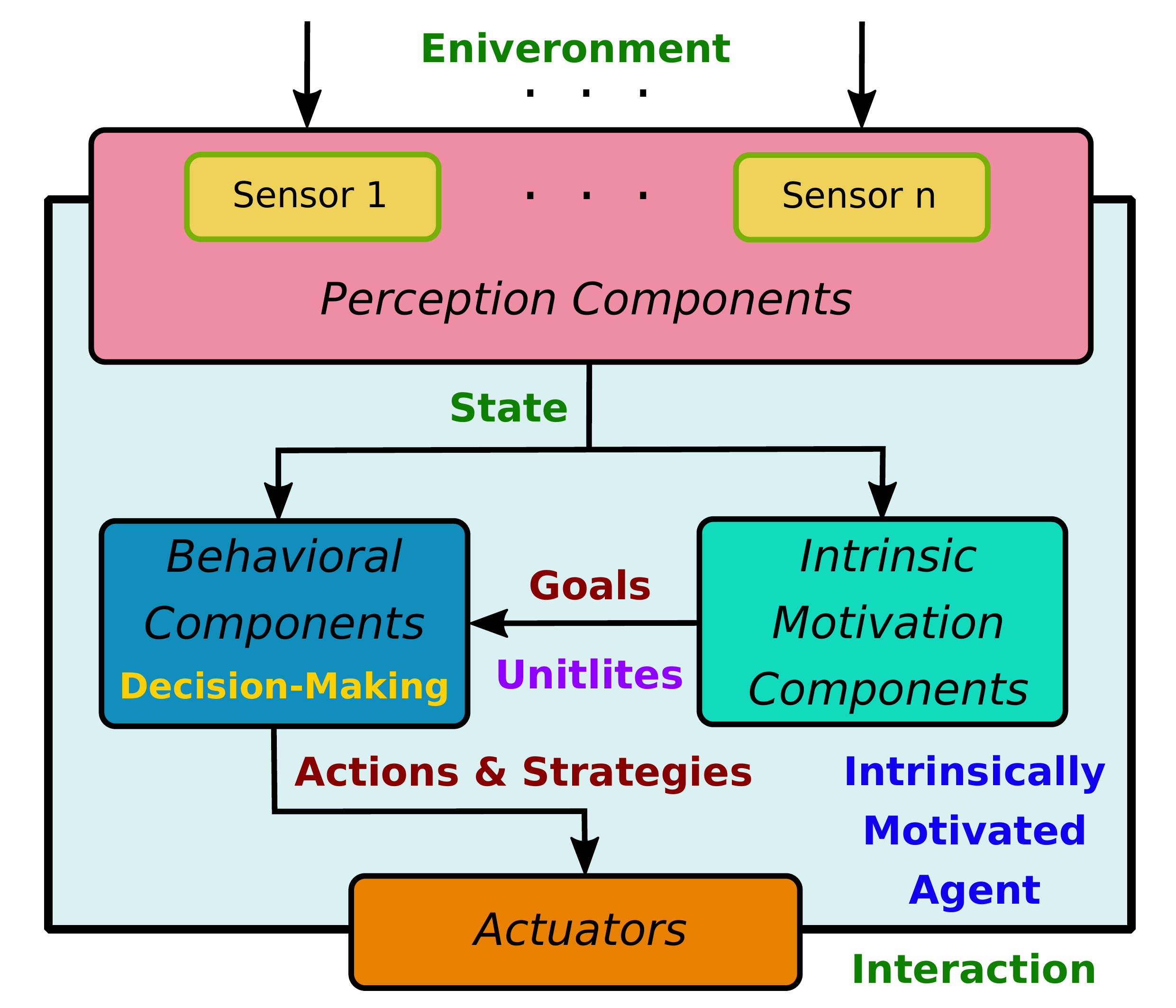}
    \caption{The illustration of the model for a generic intrinsically motivated agent.}
    \label{fig:generic_motivated}
\end{figure}

\cite{merrick2013novelty} presents a model of a generic intrinsically motivated agent, in which intrinsic motivation components modulate behavioral components to connect perception components (sensors) and actuators, as shown in Fig. \ref{fig:generic_motivated}. Through sensors generating an environment state at each time step, the agent will form various goals and corresponding expected utilities in motivation components. Then, the behavioral components will make decisions based on specific goals to modify the agent's internal state and generate diverse strategies and actions to achieve goals by triggering actuators to interact with the environment changing the external state.

\subsubsection{Needs-driven Behaviors}
In nature, from cell to human, all intelligent agents represent different kinds of hierarchical needs such as the low-level physiological needs (food and water) in microbes and animals; the high-level needs for self-actualization (creative activities) in human beings \cite{maslow1943theory}. Different levels of needs stimulate all agents to adopt diverse behaviors achieving certain goals or purposes and satisfying their various needs in the natural world \cite{yates2012self}. They also can be regarded as {\it self-organizing behaviors} \cite{banzhaf2009self}, which keep on interacting with the environments and exchanging information and energy to support the system's survival and development.

Specifically, the needs describe the necessities for a self-organizing system to survive and evolve, which arouses an agent to action toward a goal, giving purpose and direction to behavior and strategy. For example, \cite{mcfarland2018autonomy} introduced the self-sufficient robots' decision-making based on a basic work cycle -- find fuel -- refuel. They utilized some utility criterion needs, such as utility maximization, to guide robots performing opportunistic behaviors \cite{mcfarland1997basic}.

From the utility theory perspective, needs are regarded as {\it innate values} or motivations that drive agents to interact with their environments, as we discussed previously.
Especially the ultimate payoff for animals is to maximize inclusive fitness utilizing some evolutionary strategy, and for humans is to maximize profitability through some marketing strategy \cite{meyer1991animals,mcfarland1993intelligent}.
If we consider the ecological status of AI agents, like robots, accomplishing tasks in the real world, their situation is similar to our ecological situation \cite{mcfarland1994towards}.

\subsubsection{Utility-orient Needs Paradigm Systems}

For the above concepts to be cognitively applicable to the intelligent system, their biology-inspired needs are somehow associated with its actuation, which drives it to fulfill its motivations or satisfy its drivers \cite{romero2018utility}.
Moreover, before an individual agent cooperates with group members, it needs to fit some basic needs, such as having enough energy, guaranteeing other agents' safety, etc. Then it requires corresponding capabilities to collaborate with other agents, satisfy task requirements, and adapt to dynamically changing environments \cite{yang2020hierarchical}. Therefore, it is necessary to build a unified utility paradigm describing the needs of AI agents and humans in the common ground for evaluating their relationships like trust, which is also the pre-condition of safety, reliability, stability, and sustainability in cooperation \cite{yang2020needs}.
\begin{figure}
	\centering
    \includegraphics[width=\linewidth]{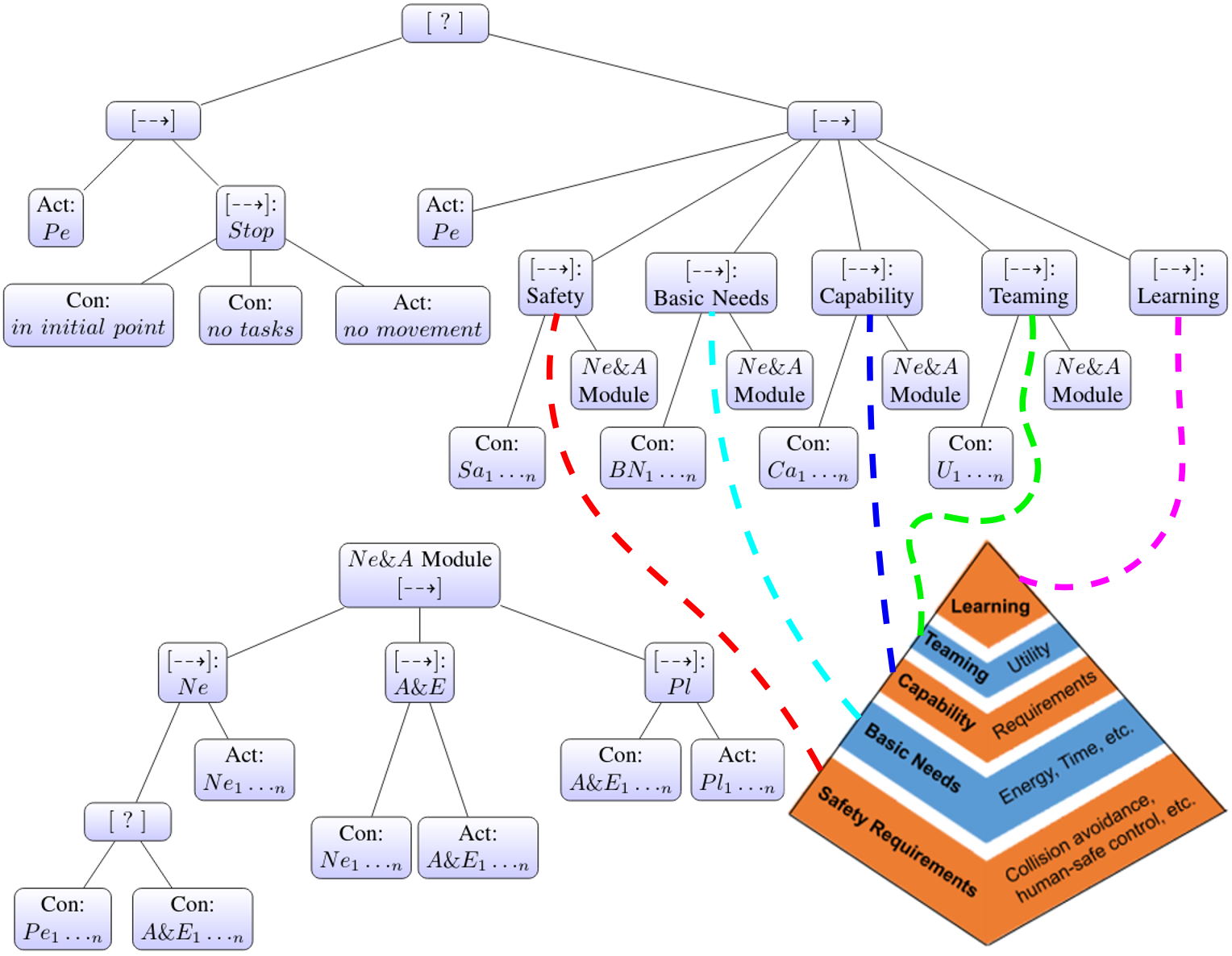}
    \caption{Behavior Tree representing \textit{Robot Needs Hierarchy} \cite{yang2020needs}. \textbf{Note*}: $[?]$ - Selector Node, $[\dashrightarrow]$ - Sequence Node, $Con$ - Conditions, $Act$ - Actions, $Pe$ - Perception, $Sa$ - Safety, $BN$ - Basic Needs, $Ca$ - Capability, $U$ - Utility, $Pl$ - Plan, $Ne$ - Negotiation, $A\&E$ - Agreement and Execution.}
    \label{fig:bt_sass}
\end{figure}
\begin{figure*}
	\centering
    \includegraphics[width=\linewidth]{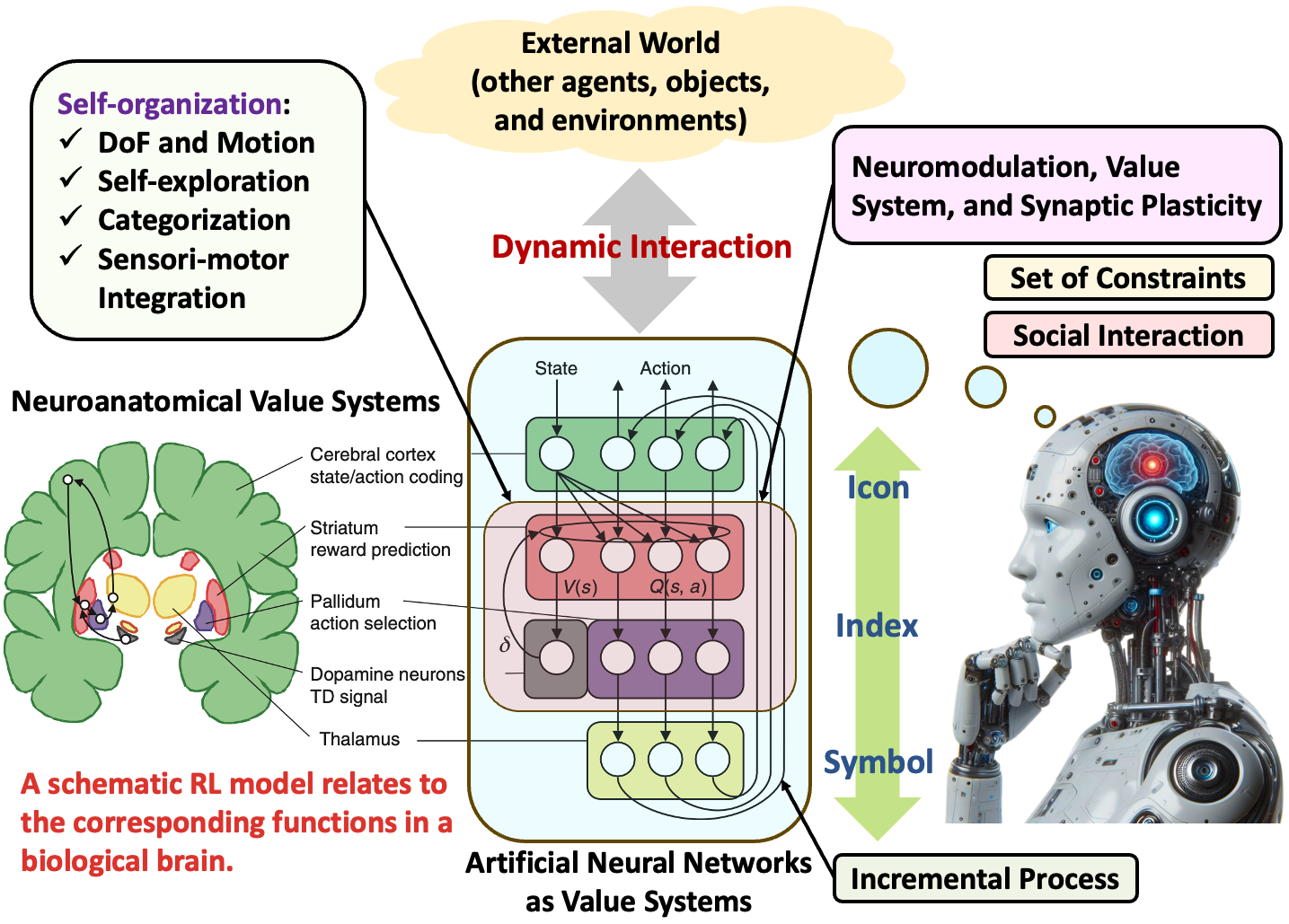}
    \caption{The illustration of the various aspects of the value system development \cite{asada2009cognitive} from a biological brain \cite{glimcher2013neuroeconomics} to an artificial brain.}
    \label{fig:RL_brain}
\end{figure*}

Similar to Maslow's hierarchy of needs, which describes humans’ motivations and value systems, we can also organize autonomous robots’ motivations as a hierarchical structure, represented as a conditional dependence network \cite{yang2021can,yang2023hierarchical}, such as a Bayesian network. The lowest (first) level is the safety features of the agent (e.g., features such as collision detection, fault detection, etc., that assure safety to the agent, human, and other friendly agents in the environment). The safety needs (Eq.~\eqref{safety_need}) can be calculated through its safety feature's value and corresponding safety feature's probability based on the current state of the agent. After satisfying safety needs, the agent considers its basic needs (Eq.~\eqref{basic_need}), which includes features such as energy levels, data communication levels that help maintain the basic operations of that agent. 
Only after fitting the safety and basic needs, an agent can consider its capability needs (Eq.~\eqref{capability_need}), which are composed of features such as its health level, computing (e.g., storage, performance), physical functionalities (e.g., resources, manipulation), etc. 

At the next higher level, the agent can identify its teaming needs (Eq.~\eqref{teaming_need}) that accounts the contributions of this agent to its team through several factors (e.g., heterogeneity, trust, actions) that the team needs so that they can form a reliable and robust team to successfully perform a given mission. 

Ultimately, at the highest level, the agent learns some skills/features to improve its capabilities and performance in achieving a given task, such as Reinforcement Learning. 
The policy features (Q table or reward function) are accounted into its learning needs expectation (Eq.~\eqref{learning_need}).
The expectation of agent needs at each level are given below:
\vspace{-2mm}
\begin{equation}
\begin{split}
    Safety~~Needs: N_{s_{j}} = \sum_{i=1}^{s_{j}} S_{i} \cdot \mathbb{P}(S_{i}|X_j, T); \label{safety_need}
\end{split}
\end{equation}
\vspace{-2mm}
\begin{equation}
\begin{split}
    Basic~~Needs: N_{b_j} = \sum_{i=1}^{b_{j}} B_{i} \cdot \mathbb{P}(B_{i}|X_j, T, N_{s_{j}}); \label{basic_need}
\end{split}
\end{equation}
\vspace{-2mm}
\begin{equation}
\begin{split}
    Capability~Needs: N_{c_j} = \sum_{i=1}^{c_{j}} C_i \cdot \mathbb{P}(C_i|X_j, T, N_{b_j}); \label{capability_need}
\end{split}
\end{equation}
\vspace{-2mm}
\begin{equation}
\begin{split}
    Teaming~~Needs: N_{t_j} = \sum_{i=1}^{t_j} T_i \cdot \mathbb{P(}T_i | X_j, T, N_{c_j}); \label{teaming_need}
\end{split}
\end{equation}
\vspace{-2mm}
\begin{equation}
\begin{split}
    Learning~~Needs: N_{l_j} = \sum_{i=1}^{l_j} L_i \cdot \mathbb{P(}L_i | X_j, T, N_{t_j}); \label{learning_need}
\end{split}
\end{equation}
{Here,} $X_j=\{P_j,C_j\}$ $\in$ $\Psi$ is the combined state of the agent $j$ with $P_j$ being the perceived information by that agent and $C_j$ representing the communicated data from other agents. $T$ is the assigned task (goal or objective). $S_i$, $B_i$, $C_i$, $T_i$, and $L_i$ are the utility values of corresponding feature/factor $i$ in the corresponding levels - Safety, Basic, Capability, Teaming, and Learning, respectively. $s_j$, $b_j$, $c_j$, $t_j$, and $l_j$ are the sizes of agent $j$'s feature space on the respective levels of needs. 

The collective need of an agent $j$ is expressed as the union of needs at all the levels in the needs hierarchy as in Eq. \eqref{eqn:need-union}. Where each level of category needs is combined with various similar needs (expected values), presenting as a set, consisting of an individual hierarchical and compound needs matrix $N_j$. Fig. \ref{fig:bt_sass} illustrates a human-like motivation mechanism in a robot, represented as a behavior tree (BT).
\begin{equation}
    N_j = N_{s_j} \cup N_{b_j} \cup N_{c_j} \cup N_{t_j} \cup N_{l_j} 
    \label{eqn:need-union}
\end{equation}

Where the needs represent motivations or goals of agents to achieve various tasks and consist of their value systems in the development.
Furthermore, when AI agents work with humans, they should satisfy humans' needs and assist humans in fulfilling their missions efficiently. Especially trust among agents can be evaluated based on their current status and needs, performance, knowledge, experience, motivations, etc. 

According to the above discussion, in the following sections, we will review the existing literature from the perspectives of single-agent systems, multi-agent systems (MAS), trust among agents, and human-robot interaction (HRI) from the utility-oriented cognitive modeling angle.

\section{Single-Agent Systems}
\label{sas}

For single-agent systems, cognitive robotics is the most typical research area for implementing {\it Utility Theory}. It studies the mechanisms, architectures, and constraints that allow lifelong and open-ended improvement of perceptual, reasoning, planning, social, knowledge acquisition, and decision-making skills in embodied machines \cite{merrick2017value}. Especially building a value system to mimic the ``brain'' of an AI agent mapping behavioral responses for sensed external phenomena is the core component of cognitive robotics, which is also an emerging and specialized sub-field in neurorobotics, robotics, and artificial cognitive systems research. 

From a neuroeconomic perspective, a variety of cortical and subcortical areas are associated with neural firing of reward-predictive and reward-prediction error signals, including the orbitofrontal cortex, prefrontal cortex, parietal cortex, and striatum.
Fig. \ref{fig:RL_brain}  illustrates a schematic model of reinforcement learning implementation in the cortico-basal ganglia circuit \cite{glimcher2013neuroeconomics} and the various aspects of value system development from a biological brain to an artificial brain, including external observation, internal structure, infrastructure, and social structure \cite{lungarella2003developmental}.
Similarly, to build a utility-based reward system in robots, existing value systems can be classified into three categories: {\it Neuroanatomical Systems} discuss the explainable biologically inspired value systems design from neuroanatomy and physiology perspectives \cite{sporns2000plasticity}; {\it Neural Networks Systems} build more abstract models through mathematical approaches to mimic the agent's value systems; {\it Motivational Systems} consider the model that agents interact with environments to satisfy their innate values, and the typical mechanism is reinforcement learning (RL). 
We review the three categories as follow:

\subsection{Neuroanatomical Systems}
In the studies of the theory of primate nervous systems in the brain, researchers usually used robotics systems as a test-bed \cite{krichmar2003brain}, figuring out how neurons influence each other, how long neurons activate, and the brain regions are affected \cite{almassy1998behavioral}. Fig. \ref{fig:brain} illustrates the functions of the brain from anatomical and physiological perspectives.
Following the goals of synthetic neural modeling, they aim to forge links between crucial anatomical and physiological properties of nervous systems and their overall function during autonomous behavior in an environment \cite{merrick2017value}.

For example, \cite{sporns2000plasticity} designed a value system in the NOMAD mobile robot, driven by the object -- ``taste'' to modulate changes in connections between visual and motor neurons, thus linking specific visual responses to appropriate motor outputs. \cite{prescott2006robot} described the robotic architecture embedding of a high-level model of the basal ganglia and related nuclei based on varying motivational and sensory inputs, such as fear and hunger, to generate coherent sequences of robot behavior.
\begin{figure}
	\centering
    \includegraphics[width=\linewidth]{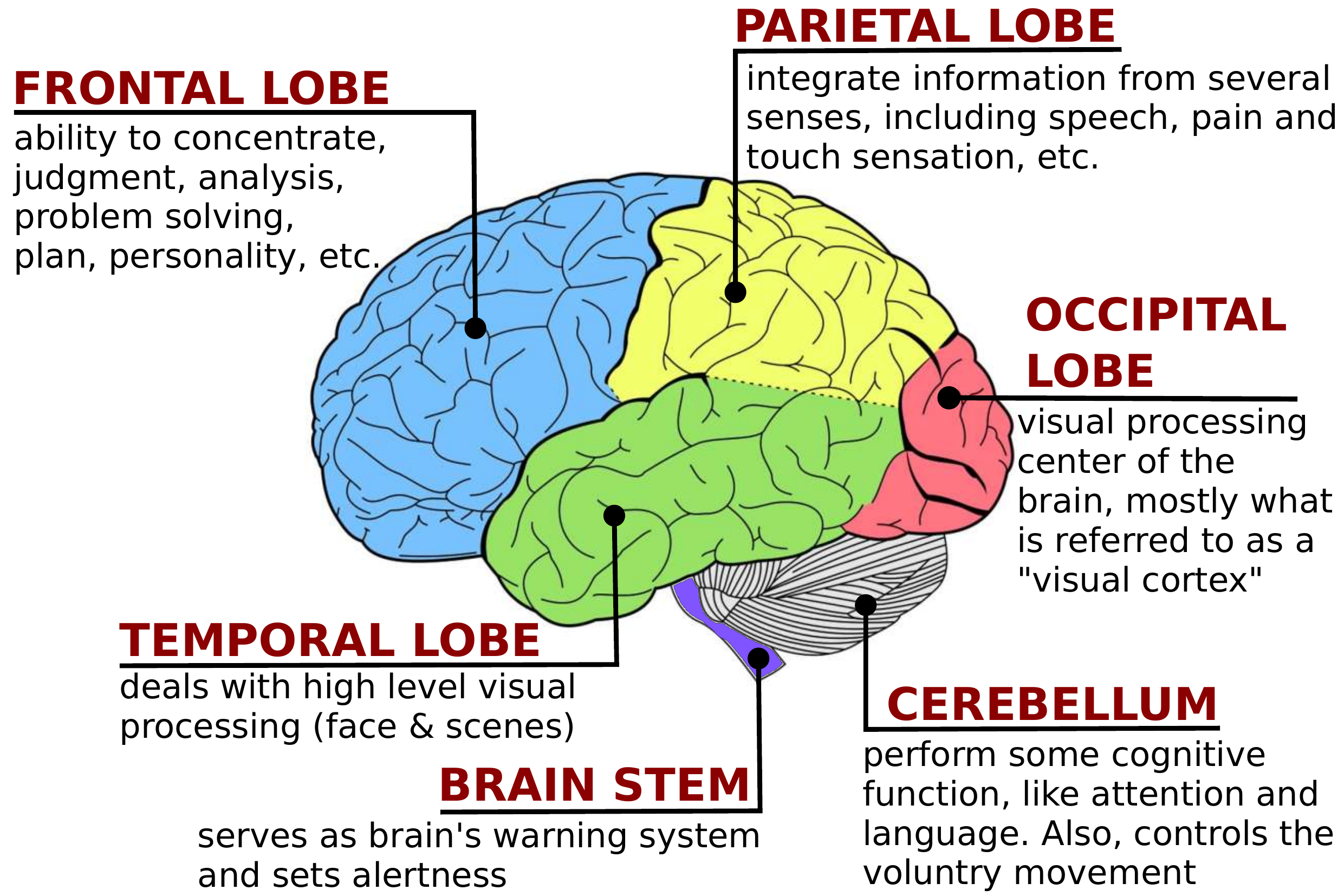}
    \caption{The illustration of the functions of the brain.}
    \label{fig:brain}
\end{figure}

Especially based on the principles of neuromodulation in the mammalian brain, \cite{cox2009neuromodulation} presents a strategy for controlling autonomous robots. They used a cognitive robot -- CARL-1, to test the hypothesis that neuromodulatory activity can shape learning, drive attention, and select actions. The experiments demonstrated that the robot could learn to approach and move away from stimuli by predicting the positive or negative value.
Moreover, \cite{fiore2014keep} shows the putative functions ascribed to dopamine (DA) emerging from the combination of a standard computational mechanism coupled with differential sensitivity to the presence of DA across the striatum through testing on the simulated humanoid robot iCub interacting with a mechatronic board. 
\begin{equation}
    \begin{split}
        \frac{dx}{dt} = -Lx + I
    \label{leaky}
    \end{split}
\end{equation}

To summarize, although researchers have proposed numerous neuroanatomical models, the most common node type is the leaky integrator neural unit. Leaky integrator units have the differential equations Eq. \eqref{leaky}.
Where $x$ is the activation potential of the neuron, $I$ is the input to the neuron, $L$ is the rate of ``leak'' of potential, and $t$ is time. Additionally, researchers use leaky integrator units to model intrinsic motivations, such as novelty, in cognitive models of value systems. It is worth mentioning that the current trend tries to blur the line between designing value systems that follow the neuroanatomy and biology of the brain and artificial neural networks (ANN) that are biologically inspired but less anatomically accurate \cite{merrick2017value}. The following section examines the ANN as a value system.

\subsection{Artificial Neural Networks Value Systems}
\label{ann}

Artificial Neural Networks (ANN) are an information manager model inspired by the biological nervous systems function of the human brain. Similarly, Fig. \ref{fig:brain} demonstrates the connection of different function modules within the human brain working as a neural network to perform various reasoning activities. Like the human brain adjusting the synaptic relationships between and among neurons in learning, ANN also needs to modify the parameters of ``nodes'' (neuron) to adapt to the specific scenario through a learning process. Fig. \ref{fig:ann} illustrates a two-layered feedforward neural network.
\begin{figure}
	\centering
    \includegraphics[width=\linewidth]{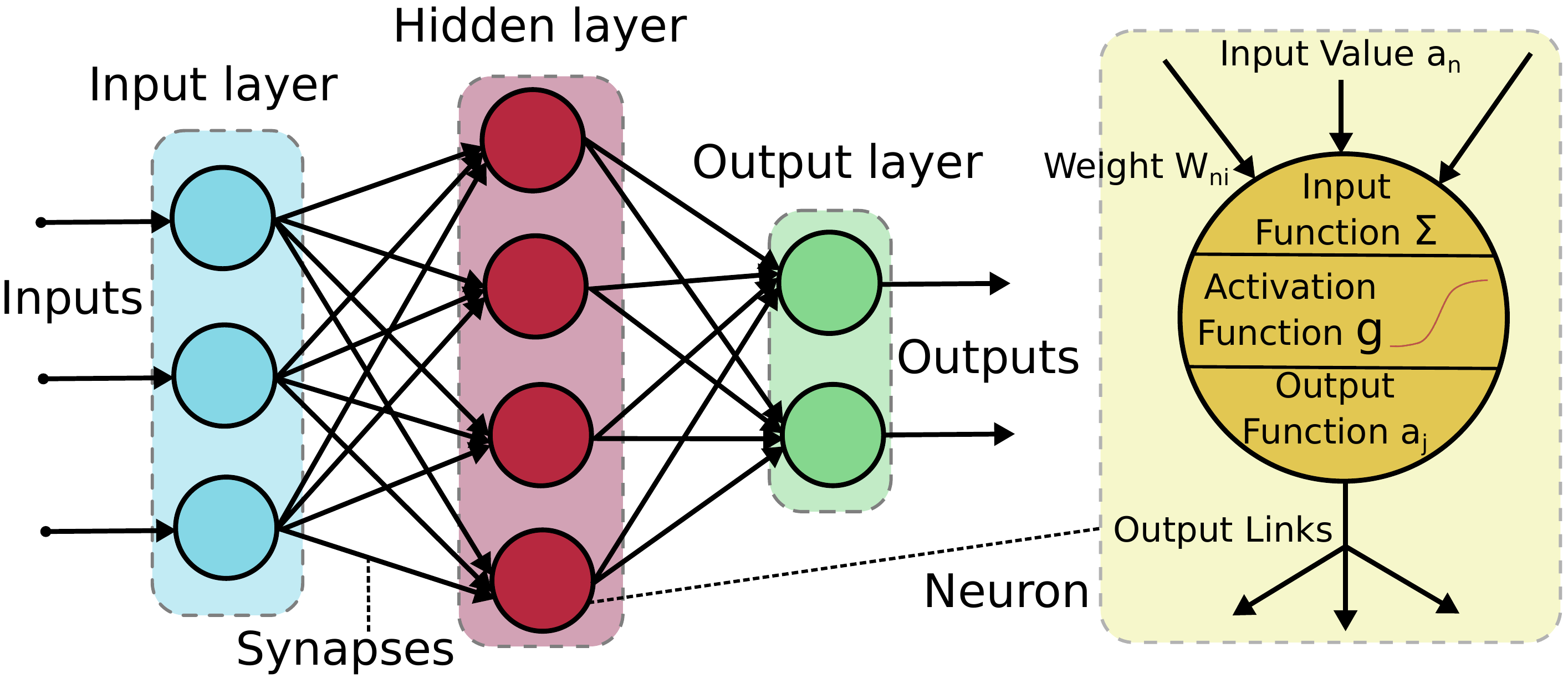}
    \caption{The illustration of a two-layered feedforward neural network.}
    \label{fig:ann}
\end{figure}

There are several typical ANN models, such as restricted Boltzmann machines (RBMs) \cite{hinton2006fast}, spiking neural networks \cite{maass1997networks}, adaptive resonance theory (ART) networks \cite{grossberg1976adaptive}, autoencoder (AE) \cite{bourlard1988auto}, convolutional neural networks (CNNs) \cite{fukushima1982neocognitron}, growing neural gases (GNGs) \cite{fritzke1994growing}, ect. In the ANN, Nodes working as neurons are non-linear processing units receiving various signals and performing different processing steps, as depicted graphically in Fig. \ref{fig:ann}. Moreover, the AI agent's sensor states represent its properties for the input nodes, and the output nodes express corresponding actions and behaviors in the current status. The agent needs to learn a function that can describe its future expected reward (utility) based on the specific action or strategy in the given state. Due to the reward inverse proportion to learning error, the agent is more motivated to learn the actions with higher learning error until it gets the skill. Then, it will switch to another to learn based on its needs.

Especially \cite{merrick2010comparative} proposes a neural network architecture for integrating different value systems with reinforcement learning, presenting an empirical evaluation and comparison of four value systems for motivating exploration by a Lego Mindstorms NXT robot. The value system is a multi-layer neural network. The first layers process sensations to transmit information to motivation layers for computing motivation values, and the reinforcement learning layers compute weight values to generate actions and update weights through Q-learning. The framework has a combined memory model for the value system and RL components, as well as computational motivation models to guide exploration and learning. Fig. \ref{fig:gnn_arch} shows the generic neural network structure used to compare different motivations in a value system.
Also, \cite{gordon2012hierarchical} introduces a model of hierarchical curiosity loops for an autonomous active learning agent by selecting the optimal action that maximizes the agent’s learning of sensory-motor correlations. 

From the information theory perspective, \cite{martius2013information,martius2014self} presents a method linking information theoretic quantities on the behavioral level (sensor values) to explicit dynamical rules on the internal level (synaptic weights) in a systematic way. They studied an intrinsic motivation system for behavioral self-exploration based on the maximization of the predictive information using a range of real robots, such as the humanoid and Stumpy robots. 
Additionally, \cite{makukhin2015exploring} proposed the value system of a developmental robot using the RBM as the data structure. They use simulation to demonstrate the mechanism allowing the agent to accumulate knowledge in an organized sequence with gradually increasing complexity while hardly learning from purely random areas. As we can see, RL has become a popular and effective model for non-specific reward tasks, especially when combined with deep learning methods implemented in motivational systems modeling, which we discuss in the following part.
\begin{figure}
	\centering
    \includegraphics[width=\linewidth]{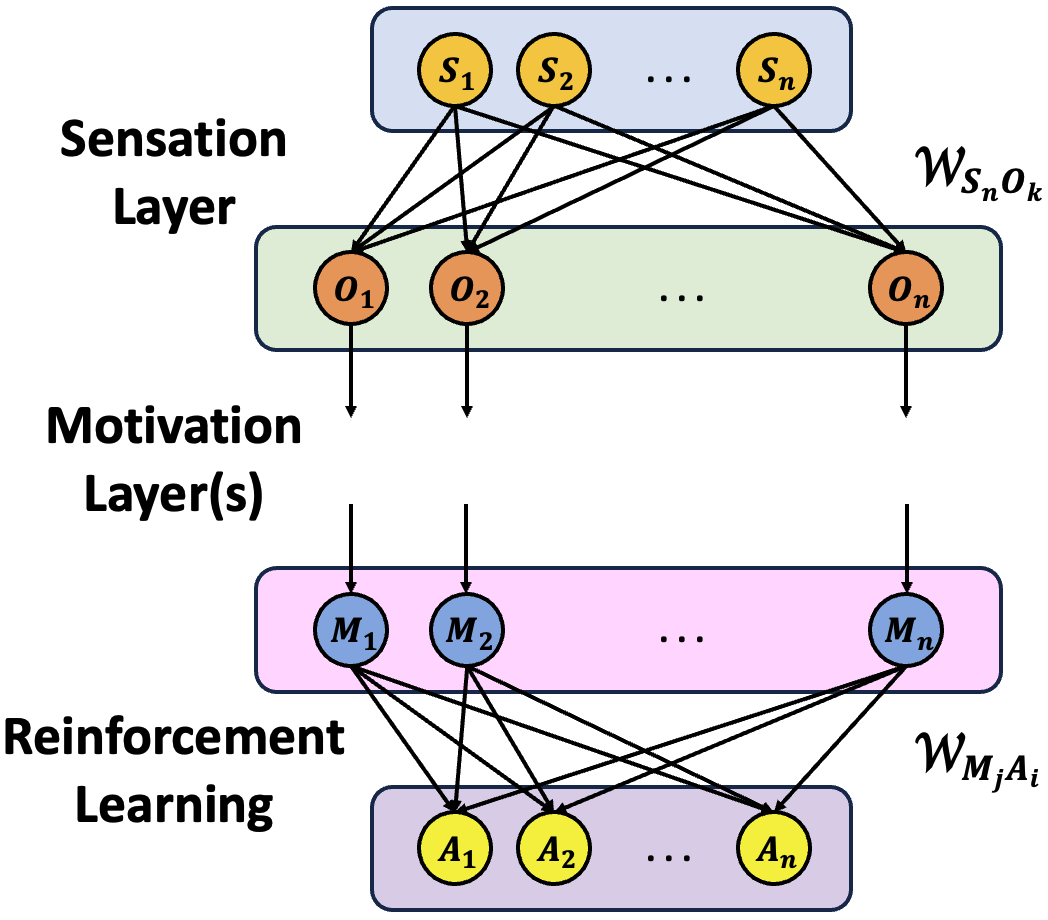}
    \caption{The illustration of the generic neural network structure used to compare different motivations in a value system \cite{merrick2010comparative}.}
    \label{fig:gnn_arch}
\end{figure}

\subsection{Motivational Systems}

To clarify agents' motivation, providing an expected utility for each perceptual state in the motivation evaluation is the pre-condition in their decision-making, especially in discovering goals achieving them automatically, and determining the priority of drives \cite{romero2019simplifying}. Many machine learning algorithms share the properties of value systems and provide a way for AI agents to improve their performance in tasks and respect their innate values \cite{merrick2017value}. 

\subsubsection{Reinforcement Learning}
\begin{figure}
	\centering
    \includegraphics[width=\linewidth]{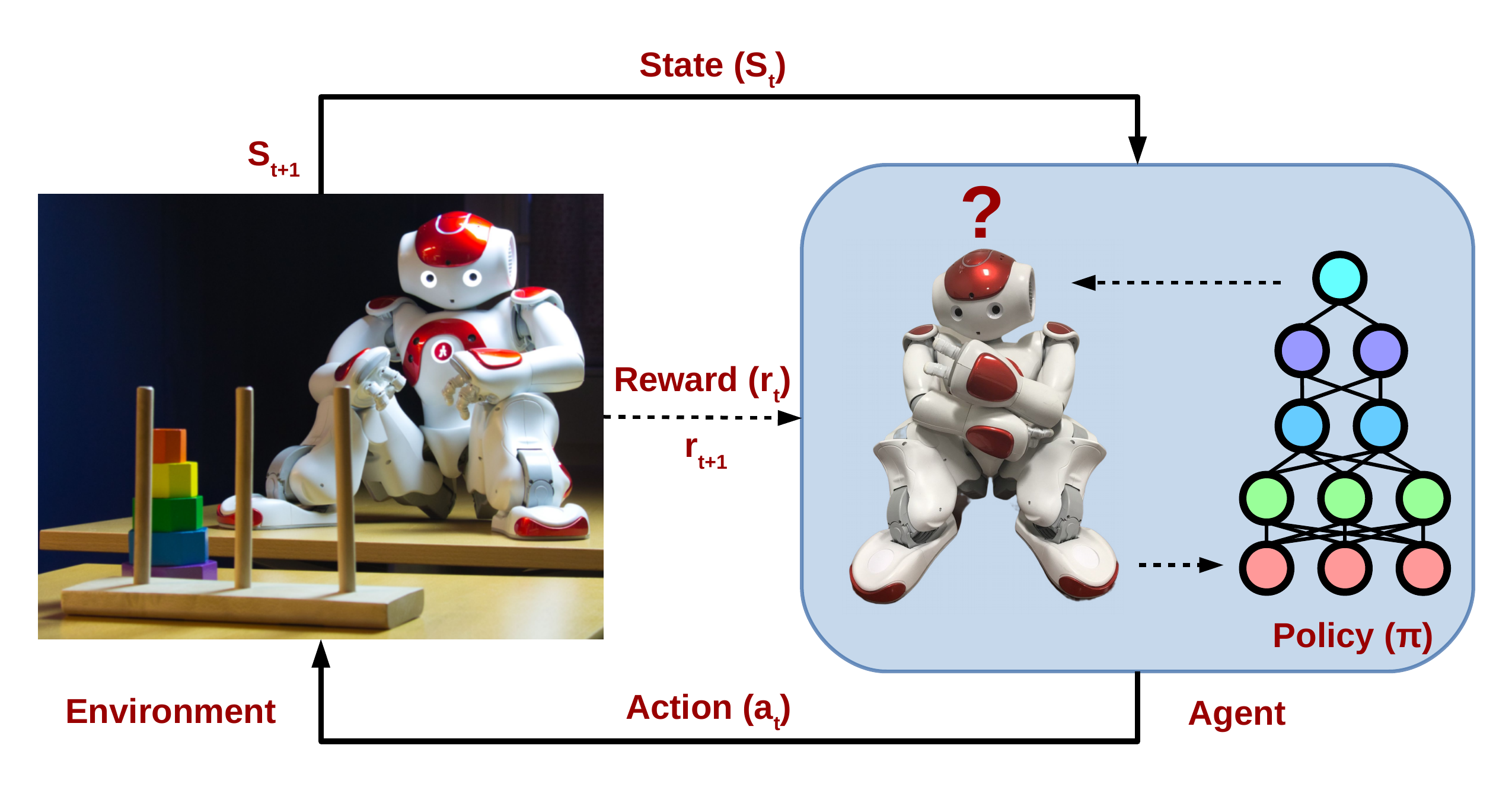}
    \caption{Rewards (Cumulative Utilities) Driven Behaviors}
    \label{fig:rewards_driven}
\end{figure}

The most typical and general model is reinforcement learning (RL). It is an excellent model to describe the needs-driven behaviors of AI agents.
The essence of RL is learning from interaction based on reward-driven (such as utilities and needs) behaviors, much like natural agents. 
When an RL agent interacts with the environment, it can observe the consequence of its actions and learn to change its behaviors based on the corresponding rewards received. 
The theoretical foundation of RL is the paradigm of trial-and-error learning rooted in behaviorist psychology \cite{sutton2018reinforcement}. In RL, the policy dictates the actions that the agent takes as a function of its state and the environment, and the goal of the agent is to learn a policy maximizing the expected cumulative rewards/utilities in the process.
Fig. \ref{fig:rewards_driven} illustrates the RL process of rewards-driven behaviors \cite{ermon2019}.

When an RL agent interacts with the environment, it can observe the consequence of its actions and learn to change its behaviors based on the corresponding rewards received. Moreover, the theoretical foundation of RL is the paradigm of trial-and-error learning rooted in behaviorist psychology \cite{sutton2018reinforcement}. In this process, the best action sequences of the agent is determined by the rewards through interaction and the goal of the agent is to learn a policy $\pi$ maximizing its {\it expected return} (cumulative utilities with discount). In a word, given a state, a policy returns an action performance and learn an optimal policy from the consequence of actions through trial and error to maximize the {\it expected return} in the environment. Figure \ref{fig:rewards_driven} illustrates this perception-action-learning loop.

\subsubsection{Integrating a Robot Utility Value System to RL}
The flexibility of RL permits it as an effective utility system to model agents' motivation through designing task-nonspecific reward signals derived from agents' experiences \cite{merrick2010comparative}. \cite{huang2002novelty} introduced one of the earliest self-motivated value systems integrated with RL, demonstrating that the SAIL robot could learn to pay attention to salient visual stimuli while neglecting unimportant input. To understand the origins of reward and affective systems, \cite{doya2005cyber} built artificial agents sharing the same intrinsic constraints as natural agents: self-preservation and self-reproduction. \cite{merrick2010modeling} uses a table-based Q-learning as a baseline compared with other function approximation techniques in motivated reinforcement learning.

From the mechanism of intelligent adaptive curiosity perspective, \cite{oudeyer2007intrinsic} introduced an intrinsic motivation system pushing a robot toward situations to maximize its learning progress. \cite{frank2014curiosity} presented a curiosity-driven reinforcement learning approach that explores the iCub's state-action space through information gain maximization, learning a world model from experience, and controlling the actual iCub hardware in real-time. To help the iCub robot intrinsically motivated to acquire, store and reuse skills, \cite{kompella2017continual} introduced Continual Curiosity driven Skill Acquisition (CCSA), a set of compact low-dimensional representations of the streams of high-dimensional visual information learned through incremental slow feature analysis. RL agents learn various skills by associating intrinsic rewards with improvements to the world model \cite{luciw2013intrinsic}. The acquired skill includes both the learned actions and the feature representation, which greatly change the feature output. Moreover, these skills will be stored and reused to generate new observations for continually developing advanced skills.
\begin{figure*}
	\centering
    \includegraphics[width=\linewidth]{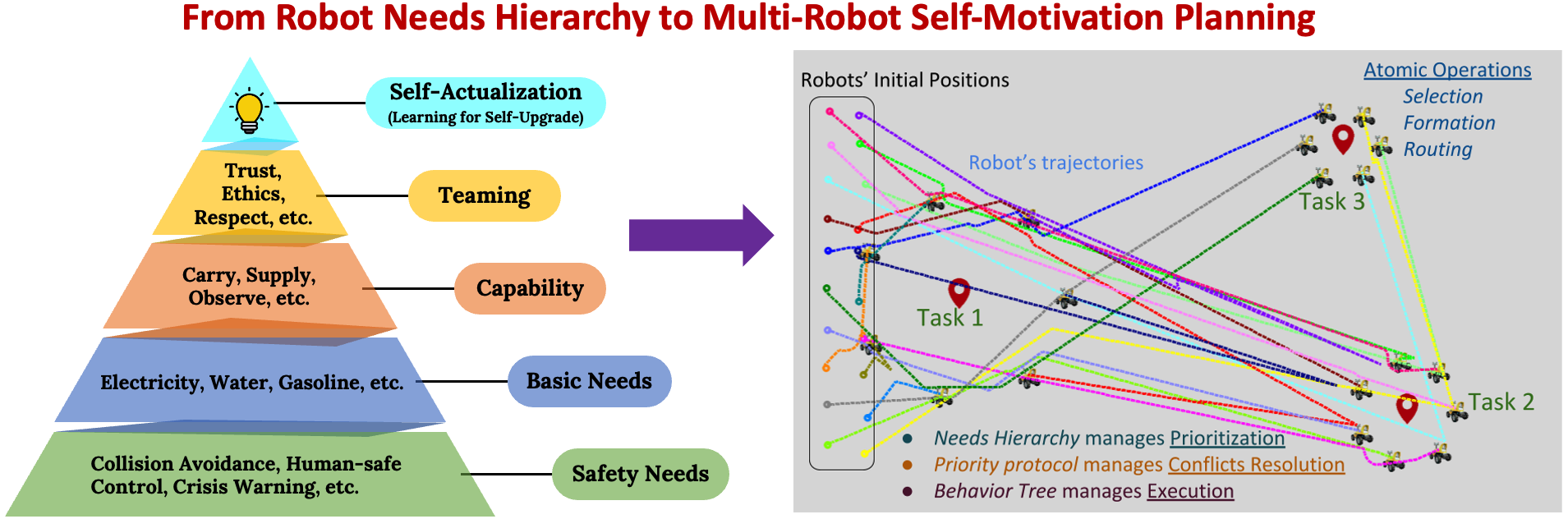}
    \caption{The illustration of {\it Robot Needs Hierarchy} as five different levels \cite{yang2020hierarchical}.}
    \label{fig:sass}
\end{figure*}

Additionally, \cite{di2014role} proposed an RL model combined with the agent basic needs satisfaction and intrinsic motivations. It was tested in a simulated survival domain, where a robot was engaged in survival tasks such as finding food or water while avoiding dangerous situations. Furthermore, \cite{yang2020hierarchical,yang2020needs} recently proposed the first formalized utility paradigm -- \textit{Agent Needs Hierarchy} -- for the needs of an AI agent to describe its motivations from the system theory and psychological perspective. The Agent Needs Hierarchy is similar to Maslow's human needs pyramid \cite{maslow1943theory}. 
An agent's abstract needs for a given task are prioritized and distributed into multiple levels, each of them preconditioned on their lower levels. At each level, it expresses the needs as an expectation over the corresponding factors/features' distribution to the specific group \cite{yang2020needs}. Ultimately, at the highest level, the agent learns skills or features to improve its capabilities and performance in achieving a given task through RL. 
Fig. \ref{fig:sass} illustrates the {\it Agent Needs Hierarchy} as five different levels in the proposed Self-Adaptive Swarm Systems (SASS). 

Later work by \cite{10.1145/3605098.3636113,yang2024bayesian} presented a strategy-oriented Bayesian soft actor-critic (BSAC) Model based on the agent needs, which integrates an agent strategy composition approach termed Bayesian Strategy Network (BSN) with the soft actor-critic (SAC) method to achieve efficient deep reinforcement learning (DRL). The key idea of the BSAC is that by building a suitable BSN to describe complicated relationships among subsystems in the complex system, it can transform a complex policy into a joint policy consisting of several simple sub-policies, which can improve the sample efficiency and training speed. From another perspective, the subsystems' policy training is like muscle memory training in human skills development. Fig. \ref{fig:bsac} illustrates a Bayesian Strategy Network for a robot dog model that implements the Actor-Critic DRL architecture.

Furthremore, \cite{yang2025innate} propose a new RL model termed innate-values-driven RL (IVRL) based on combined motivations' models and expected utility theory to mimic its complex behaviors in the evolution through decision-making and learning. 
Compared with the traditional RL model, IVRL generates the input state and rewards from the critic module instead of directly from the environment, which means that the AI agent receives various utilities from the environment through executing an action or strategy in the IVRL model. Moreover, the individual needs to calculate innate values (expected utility) through its needs weights and current utilities and then select suitable actions or strategies to optimize or maximize its accumulated expected utility. Fig. \ref{fig:single_ivrl} illustrates the IVRL model. 
By adjusting needs weights in its innate-values system, it can adapt to different tasks representing corresponding characteristics to maximize the rewards efficiently (Eq. \eqref{q_star}). In other words, through interacting with other agents and environments, the agent builds its unique value system to adapt to them, the same as "Individual is the product of their own environment." Generally, the set of agent needs reflects its innate values and motivations. It is described as the union of needs at all levels in the needs hierarchy, forming the individual hierarchical and compound needs matrix, which drives the agent to develop various strategies and diverse skills to satisfy its intrinsic needs and values, much like a human does \cite{yang2022self}. 
\begin{equation}
\begin{split}
Q^*(s, w, a) & = \max_{\omega, \pi} \mathbb{E}[G_t|S_t = s_t, W_t = w_t, A_t = a_t, \omega, \pi] \\
& = \mathbb{E}_{s' \sim \epsilon}\left[r + \gamma \max_{w', a'}Q^*(s', w', a')\bigg|s, w, a\right]
\label{q_star}
\end{split}
\end{equation}



\begin{figure*}
	\centering
    \includegraphics[width=\linewidth]{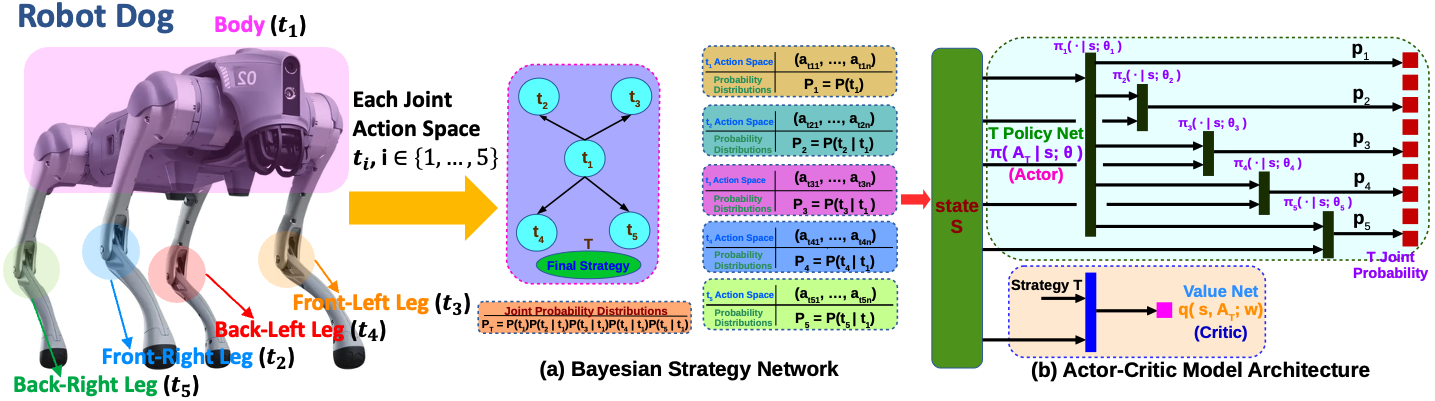}
    \caption{An Bayesian Strategy Network model of a robot dog implementing the Actor-Critic DRL architecture model \cite{10.1145/3605098.3636113}.}
    \label{fig:bsac}
\end{figure*}

To summarize, when integrating value systems into RL, we consider model-based and model-free learning alongside exemplar and function approximation learning. Two commonly used value updates modify either a utility value associated with the expected future reward of being in a given state (Eq. \eqref{motivation_value_function1}) or a utility value associated with the expected future reward of taking a given action in a given state (Eq. \eqref{motivation_value_function2}) \cite{merrick2017value}. Where the value updates involve multiple utility reward types, including intrinsic (novelty, curiosity, creativity, and competence/learning progress) and extrinsic (fear, thirst, and hunger).
\begin{equation}
    \begin{split}
        V_t(S_{t-1}) = & V_{t-1}(S_{t-1}) + \\
        & \alpha \left[R_t + \gamma V_{t-1}(S_t) - V_{t-1}(S_{t-1}) \right];
    \label{motivation_value_function1}
    \end{split}
\end{equation}
\begin{align}
    \begin{split}
        & Q_t(S_{t-1} - A_{t-1}) = Q_{t-1}(S_{t-1} - A_{t-1}) +  \\
        & \alpha \left[R_t + \gamma \max_{A \in \mathcal{A}} Q_{t-1}(S_t - A) - Q_{t-1}(S_{t-1} - A_{t-1}) \right]
    \label{motivation_value_function2}
    \end{split}
\end{align}

\section{Multi-Agent Systems}
\label{mas}

A multi-agent system (MAS) consists of several agents interacting with each other, which may act cooperatively, competitively, or exhibit a mixture of these behaviors \cite{vlassis2007concise}. In the most general case, agents will act on behalf of users with different goals and motivations to cooperate, coordinate, and negotiate with each other, achieving successful interaction, such as in human society \cite{wooldridge2009introduction}. Most MAS implementations aim to optimize the system's policies with respect to individual needs and intrinsic values, even though many real-world problems are inherently multi-objective \cite{ruadulescu2020multi}. Therefore, many conflicts and complex trade-offs in the MAS need to be managed, and compromises among agents should be based on the utility mapping the innate values of a compromise solution -- how to measure and what to optimize \cite{zintgraf2015quality}. 

However, in the MAS setting, the situations will become much more complex when we consider individual utility reflects its own needs and preferences. For example, although we assume each group member receives the same team rewards in fully cooperative MAS, the benefits received by an individual agent are usually significant differences according to its contributions and innate values in real-world scenarios or general multi-agent settings. Especially in distributed intelligent systems, such as multi-robot systems (MRS), self-driving cars, and delivery drones, there is no single decision-maker with full information and authority. Instead, the system performance greatly depends on the decisions made by interacting entities with local information and limited communication capabilities \cite{paccagnan2022utility}. 

From the multi-objective multi-agent perspective, \cite{ruadulescu2020multi} proposes a taxonomy based on utility into three types: {\it Individual utility}, where each agent serves a different agenda and just tries to optimise for that;  {\it Team utility}, where all agents serve the same interest, like playing football; {\it Social choice utility}, which represents agents are interested in optimizing the overall social welfare across all agents. According to different application scenarios, they provided a more detailed classification. Fig. \ref{fig:social_utility} illustrates the taxonomy of multi-objective multi-agent decision-making settings based on utility theory and Fig. \ref{fig:multi_utility} presents the utility rewards classification in multi-objective multi-agent decision-making with different setting.

This section, in combination with the applications in AI and robotics, reviews work on decision-makers' compromise in MAS, ranging from game-theoretic utilities to cooperative decision-making grounded in their needs and innate values. Then, we discuss the design and the influence of individual utility functions on desirable system behaviors and optimization criteria. Finally, we extend the topic from individual utility to applications of artificial social welfare.

\subsection{Game-theoretic Utility Systems}
\begin{figure*}
\centering
\includegraphics[width=2.05\columnwidth]{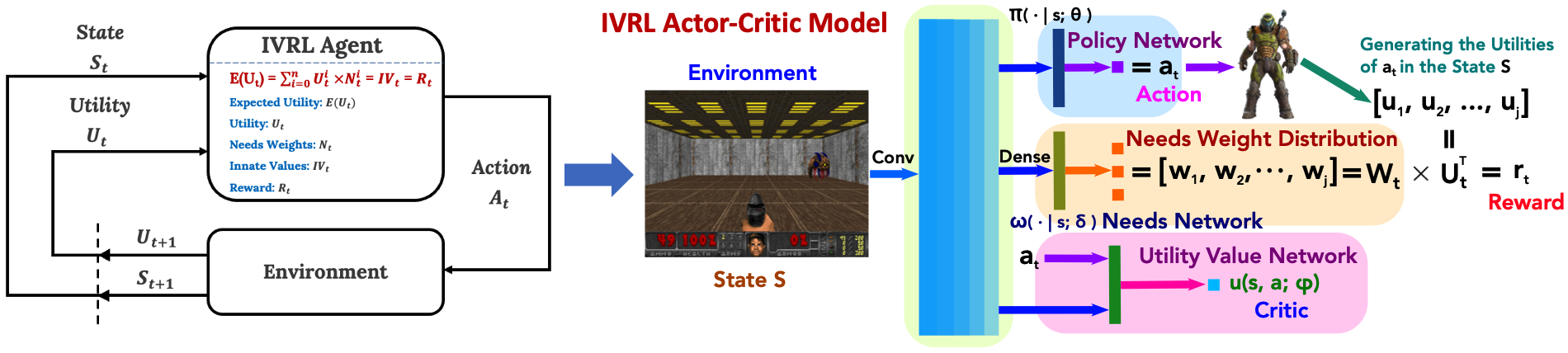}
\caption{The illustration of the architecture of the IVRL Actor-Critic model \cite{yang2025innate}.}
\label{fig:single_ivrl}
\end{figure*}

In multi-objective decision-making, each agent aims to optimise its reward utility across a set of objectives and to make compromises between competing objectives based on its utility function. However, when multiple self-interested agents learn and act together in the same environment, it is generally not possible for all agents to receive the maximum possible reward. Therefore, MAS are often designed to converge to a Nash equilibrium \cite{daskalakis2009complexity}.

From a game-theoretic control perspective, most of the research in game theory has been focused on single-stage games with fixed, known agent utilities \cite{browning2005stp}, such as distributed control in communication \cite{zhang2014game} and task allocation \cite{bakolas2021decentralized}. Especially, recent MAS research domains focus on solving path planning problems for avoiding static or dynamical obstacles \cite{agmon2011multi} and formation control \cite{shapira2015path} from the unintentional adversary perspective. For intentional adversaries, the {\it Pursuit Domain} \cite{chung2011search,kothari2014cooperative} primarily deals with how to guide pursuers to catch evaders \cite{makkapati2019optimal} efficiently. Similarly, the robot soccer domain \cite{nadarajah2013survey} deals with how one group of robots wins over another group of robots on a typical game element. Fig~\ref{fig:adversarial_scenario} illustrates a group of Unmanned Aerial Vehicles (UAV) working in adversarial environments with several typical scenarios.

Furthermore, optimal control of MAS via game theory assumes a system-level objective is given, and the utility functions for individual agents are designed to convert a Multi-Agent System into a potential game \cite{liu2019game}. Although game-theoretic approaches has received intensive attention in the control community, it is still a promising new approach to distributed control of MAS. Especially dynamic non-cooperative game theory using distributed optimization in MAS has been studied in \cite{bacsar1998dynamic}. In \cite{marden2009cooperative}, the authors investigated the MAS consensus and the work in \cite{xu2018nash} provided an algorithm for large-scale MAS optimization. However, designing utility functions, learning from global goals for potential game-based optimization of control systems, and converting the local optimization problem to an original optimization problem into a potential networked game are still open challenges \cite{abouheaf2014multi}.

It is worth mentioning that \cite{gopalakrishnan2011architectural} and \cite{marden2017game} introduced an architectural overview from the perspectives of decomposing the problem into utility function design and building the learning component. Specifically, the utility function design concerns the satisfaction of the results of emergent behaviors from the standpoint of the system designer, and the learning component design focuses on developing distributed algorithms coordinating agents towards one such equilibrium configuration. Recently, \cite{paccagnan2019utility,paccagnan2021utility} solved the utility function design of the separable resource allocation problems through systematic methodology optimizing the price of anarchy. \cite{marden2018game} discussed the learning in games of rational agents' coverage to an equilibrium allocation by revising their choices over time in a dynamic process. 

More recently, \cite{yang2023hierarchical,yang2022game} proposed a new model called the Game-Theoretic Utility Tree (GUT), combining the core principles of game theory and utility theory to achieve cooperative decision-making for MAS in adversarial environments. It combines with a new payoff measure based on agent needs for real-time strategy games. By calculating the Game-Theoretic Utility Computation Unit distributed at each level, the individual can decompose high-level strategies into executable low-level actions. It demonstrated the GUT in the proposed Explore Game domain and verified the effectiveness of the GUT in the real robot testbed -- Robotarium \cite{pickem2017robotarium}, indicating that GUT can organize more complex relationships among MAS cooperation and help the group achieve challenging tasks with lower costs and higher winning rates. Fig. \ref{fig:gut} demonstrates five pursuers catching one evader in the Pursuit-Evasion game through the game-theoretic utility tree (GUT) with circle formation (level 1) and PP tactic (level 2) in Robotarium.

\subsection{Cooperative Decision-making and Learning}
\begin{figure}
	\centering
    \includegraphics[width=\linewidth]{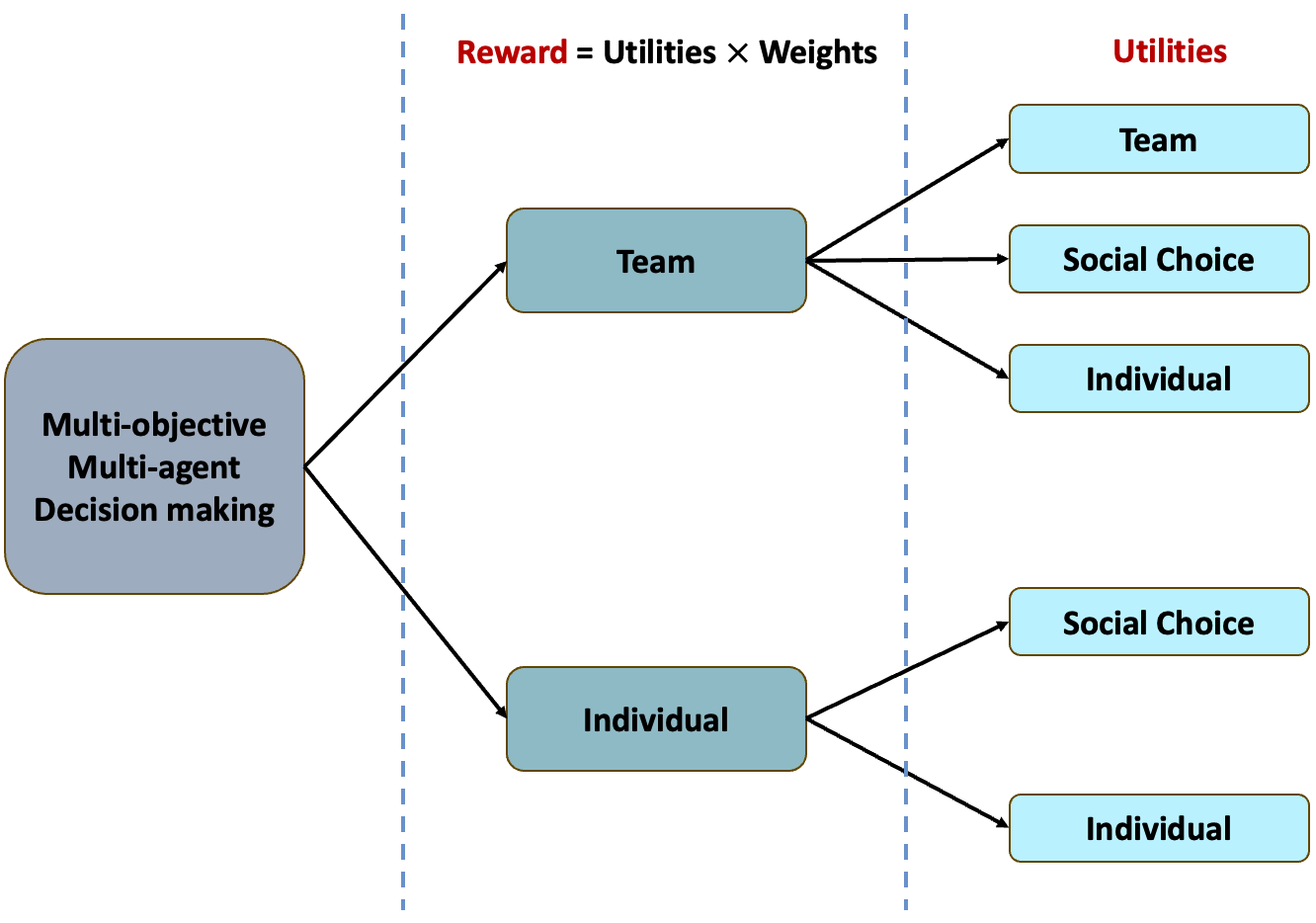}
    \caption{The illustration of the taxonomy of multi-objective multi-agent decision making settings \cite{ruadulescu2020multi}.}
    \label{fig:social_utility}
\end{figure}

Cooperative decision-making among the agents is essential to address the threats posed by intentional physical adversaries or determine tradeoffs in tactical networks \cite{cho2014tradeoffs}.  
Although cooperative MAS decision-making used to be studied in many separate communities, such as evolutionary computation, complex systems, game theory, graph theory, and control theory \cite{rizk2018decision}, these problems are either be episodic or sequential \cite{russell2002artificial}. Agents' actions or behavior are usually generated from a sequence of actions or policies, and the decision-making algorithms are evaluated based on utility-orient policy optimality, search completeness, time complexity, and space complexity.

Furthermore, existing cooperative decision-making models use the Markov decision process (MDP) and its variants \cite{pmlr-v97-yu19e}, game-theoretic methods, and swarm intelligence \cite{rizk2018decision}. They mostly involve using Reinforcement Learning (RL) and Recurrent Neural Networks (RNN) to find optimal or suboptimal action sequences based on current and previous states for achieving independent or transferred learning of decision-making policies \cite{zhang2019collaborative,pmlr-v80-rashid18a}.
Considering decision-making in the context of cooperative MAS, the learning process can be centralized or decentralized. \cite{panait2005cooperative} divides it into two categories: {\it team learning} and {\it concurrent learning}.

In \textit{team learning}, only one learner involves the learning process and represents a set of behaviors for the group of agents, which is a simple approach for cooperative MAS learning since the standard single-agent machine learning techniques can handle it. 
So considering multiple learning processes improve the team's performance, \textit{concurrent learning} is the most common approach in cooperative MAS. Since concurrent learning projects the large joint team search space onto separate, smaller subset search spaces, \cite{jansen2003exploring} argue that it is suitable for the domains that can be decomposed into independent problems and benefit from them. Especially for individual behaviors that are relatively disjoint, it can dramatically reduce search space and computation complexity. Furthermore, breaking the learning process into smaller chunks provides more flexibility for the individual learning process using computational resources.

However, \cite{panait2005cooperative} argue that the central challenge for concurrent learning is that each learner is adapting its behaviors in the context of other co-adapting learners over which it has no control, and there are three main thrusts in the area of concurrent learning: {\it credit assignment}, {\it the dynamics of learning}, and {\it teammate modeling}.
      
The {\it credit assignment} problem focuses on appropriately apportioning the group rewards (utilities) to the individual learner. The most typical solution is to separate the team rewards for each learner equally or the reward changing trend of individual learners are the same. The approach divvying up among each learner and receiving rewards through joint actions or strategies is usually termed {\it global rewards}. \cite{wolpert2002optimal} argue that global reward does not scale well to increasingly difficult problems because the learners do not have sufficient feedback tailored to their own specific actions.
      
In contrast, if we do not divide the team rewards equally, evaluating each agent's performance is based on individual behaviors, which means agents do not have the motivation to cooperate and tend to greedy behaviors; these methods are the so-called {\it local reward}. Through studies of different credit assignment policies, \cite{balch1997learning,Balch99} argue that local reward leads to faster learning rates, but not necessarily to better results than global reward. Because local rewards increase the homogeneity of the learning group, he suggests that credit assignment selection relies on the desired degree of specialization. Furthermore, \cite{mataric1994learning} argues that individual learning processes can be improved by combining separate local reinforcement with types of social reinforcement. 
\begin{figure*}
	\centering
    \includegraphics[width=\linewidth]{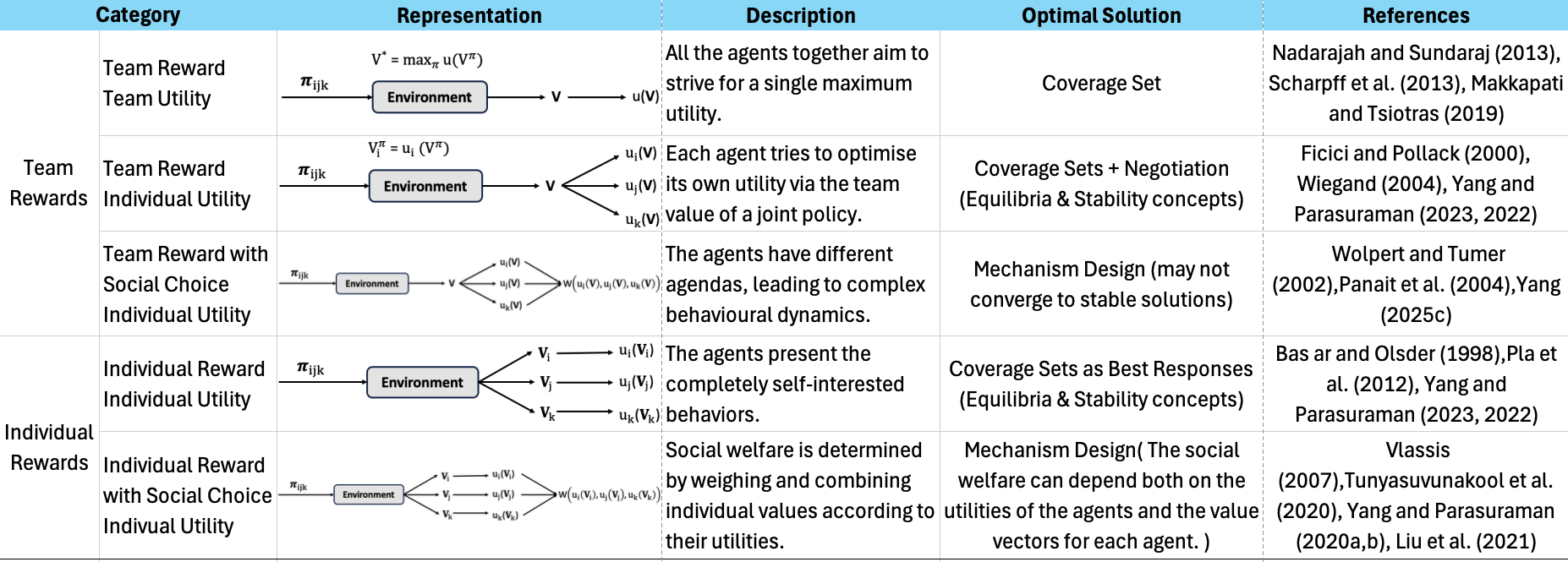}
    \caption{The utility rewards classification in multi-objective multi-agent decision-making with different setting \cite{ruadulescu2020multi}.}
    \label{fig:multi_utility}
\end{figure*}

{\it The dynamics of learning} consider the impact of co-adaptation on the learning processes. Assuming agents work in dynamic changing and unstructured environments, they need to constantly track the shifting optimal behavior or strategy adapting to various situations, especially considering the agents may change other group members' learning environments. \cite{esteban2017model} proposes a method using probabilistic and learning methods for social robots' decision-making based on affective generation. Evolutionary game theory provides a common framework for cooperative MAS learning. \cite{ficici2000game,wiegand2004analysis} studied the properties of cooperative coevolution and \cite{panait2004visual} visualize {\it basins of attraction} to Nash equilibria for cooperative coevolution. More recently, \cite{11079504} introduces a general MAS innate-values reinforcement learning architecture from the perspective of individual preferences.

Moreover, lots of research in concurrent learning involve game theory to investigate MAS problems \cite{fudenberg1998theory,shoham2008multiagent}. Especially for Nash equilibria, it provides a joint strategy for each group member, which prevents individuals from being motivated to shift their strategy or action in the current situation in terms of a better reward. In the {\it full-cooperative} scenario with global reward, the reward affects the benefits of each agent and only needs to consider the globally optimal Nash equilibrium. However, the agents' relationships are unclear for more general situations, which combine the cooperative and non-cooperative scenarios. In other words, individual rewards do not directly relate to the team reward, so the {\it general-sum game} \cite{shoham2008multiagent} is the cooperative learning paradigm.

\subsection{From Individual Utility to Social Welfare}

As we discussed above, the reward same as the utility, reflects the agents' diverse intrinsic motivations and needs. Individual and team rewards also directly describe the relationships between each group member and the team. 
We are taking MAS football and the Robot-soccer domain as an example. When considering multi-agent systems (MAS) cooperation as a team, an individual agent must first master essential skills to satisfy low-level needs (safety, basic, and capability needs). Then, it will develop effective strategies fitting middle-level needs (like teaming and collaboration) to guarantee the systems' utilities. Through learning from interaction, MAS can optimize group behaviors and present complex strategies adapting to various scenarios and achieving the highest-level needs, and fulfilling evolution. By cooperating to achieve a specific task, gaining {\it expected utility (reward)}, or against the adversaries decreasing the threat, intelligent agents can benefit the entire group development or utilities and guarantee individual needs and interests.
\cite{liu2021motor} developed the end-to-end learning implemented in the MuJoCo multi-agent soccer environment, which combines low-level imitation learning, mid-level, and high-level reinforcement learning, using transferable representations of behavior for decision-making at different levels of abstraction. 
Figure \ref{fig:humanoid_football} represents the training process.

However, suppose we extend the scalar to the entire system, like community and society, and consider optimizing the overall social welfare across all agents. In that case, we have to introduce the {\it social choice utility}. In this setting, agents have different agendas driven by their needs, leading to complex dynamic behaviors when interacting, which causes the utility hard to predict, let alone optimize. To solve this problem, we usually assume that agents are self-interested and optimize their utilities from the perspective of socially favorable desires and needs \cite{ruadulescu2020multi}. By defining the social choice function representing social welfare reflecting each agent's needs and desired outcome, we can design a system of payments making the joint policy converge to the desired outcome \cite{vlassis2007concise}. Moreover, \cite{scharpff2013coordinating} discussed computing a convex coverage set for a cooperative multi-objective multi-agent MDP balancing traffic delays and costs as a posteriori in traffic network maintenance planning.

From another perspective, considering individual rewards and utilities from the standpoint of social choice, they are weighed and summed up, consisting of a social welfare function. In other words, the social welfare function relies on individual values or return rewards and its utility computing model or mechanism. \cite{pla2012multi} discussed which properties must satisfy a multi-criteria function so it can be used to determine the winner of a multi-attribute auction. Here, social welfare may depend on attributes of the winning bids, as well as a fair outcome in terms of payments to the individual agents, that together with the costs the agents need to support to execute their bids, if chosen, typically determine the individual utilities \cite{ruadulescu2020multi}.

Generally speaking, the design of the social welfare function in MAS aims to find a mechanism for making agents' utility function transparently reflect their innate values and needs, which can optimize the joint policy concerning sustainable social welfare and guarantee individual development.

\section{Trust among Agents}
\label{taa}

Trust describes the interdependent relationship between agents \cite{swinth1967establishment}, which can help us better understand the dynamics of cooperation and competition, the resolution of conflicts, and the facilitation of economic exchange \cite{lewicki2006models}. From the economic angle, \cite{arai2009defining} has regarded trust as an expectation or a subjective probability and defined it using expected utility theory combined with concepts such as betrayal premium.

In \textit{Automation}, authors in \cite{lee2004trust} describe trust in human-automation relationships as the attitude that an agent will help achieve an individual’s goals in a situation characterized by uncertainty and vulnerability. The trustor here is a human and the trustee usually is the automation system or an intelligent agent like robots. Those systems' primary purpose is to assess and calibrate the trustors' trust beliefs reflecting the automation system's ability to achieve a specific task.

In \textit{Computing and Networking}, the concept of trust involves many research fields, such as artificial intelligence (AI), human-machine interaction, and communication networks \cite{cho2010survey,sherchan2013survey,wang2020survey}. They regard an agent's trust as a subjective belief that represents the reliability of the other agents' behaviors in a particular situation with potential risks. Their decisions are based on learning from the experience to maximize its interest (or utility) and minimize risk \cite{cho2015survey}. Especially in the social IoT domain \cite{chen2015trust,mohammadi2019trust}, trust is measured based on the information accuracy and agents' intentions (friendly or malicious) according to the current environment to avoid attacks on the system. \cite{quercia2007mate} proposes a decision-making model named MATE, which applies the expected-utility decision rule to decide whether or not pervasive devices are in their node interests to load a particular component.
\begin{figure}
	\centering
    \includegraphics[width=\linewidth]{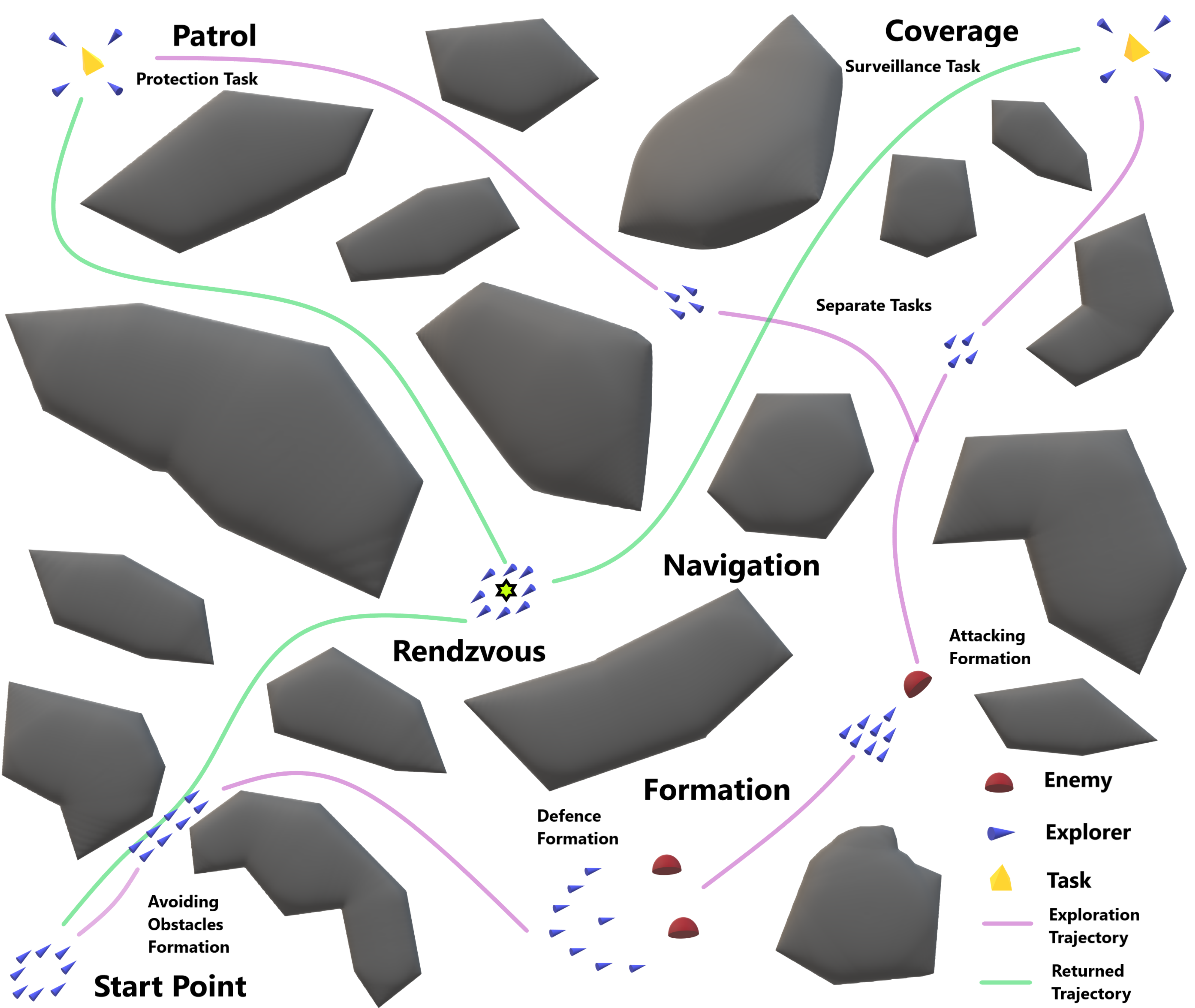}
    \caption{Illustration of The MAS (UAV) Working in Adversarial Environments with Several Typical Scenarios}
    \label{fig:adversarial_scenario}
\end{figure}

In \textit{MAS}, trust is defined by mainly two major ways. One way is reliable relations between human operators and the MAS/MRS. In this context, trust reflects human expectations and team performance assessments. The primary research includes using the trust to measure the satisfactory degree of robot team performance that hybrid considering in-process behaviors and final mission accomplishments. In this process, the visible and invisible behaviors regarding humans supervising robots and robots' behaviors will be investigated to decide the most influential factors. Especially investigating human trust in robots will help the robots to make decisions or react to emergencies, such as obstacles, when human supervision is unavailable.

The second type of trust will be defined as reliable relations among agents, reflecting the trusting agents' mission satisfaction, organizational group member behaviors and consensus, and task performance at the individual level. Moreover, trust has been associated with system-level analysis, such as reward and function at the organized group structure in a specific situation, the conflict solving of heterogeneous and diverse needs, and the system evolution through learning from interaction and adaptation. 

On the other hand, the act of trusting by an agent can be conceptualized as consisting of two main steps: 1) \textit{trust evaluation}, in which the agent assesses the trustworthiness of potential interaction partners; 2) \textit{trust-aware decision-making}, in which the agent selects interaction partners based on their trust values \cite{yu2013survey}. Also, assessing the performance of agent trust models is of concern to agent trust research \cite{fullam2005specification}.

It is worth noticing that reputation and trust mechanisms have gradually become critical elements in MAS design since agents depend on them to evaluate the behavior of potential partners in open environments \cite{pinyol2013computational}. As an essential component of reputation, trust generates the reputation from agents' previous behaviors and experiences, which becomes an implicit social control artifact \cite{conte2002reputation} when agents select partners or exclude them due to social rejection (bad reputation). Fig. \ref{fig:trust} illustrates the classification of trust modeling approaches from individual to multi-agent systems from different perspectives.

Furthermore, considering the interactions between human agents and artificial intelligence agents like human-robot interaction (HRI), building stable and reliable relationships is of utmost importance in MRS cooperation, especially in adversarial environments and rescue missions \cite{yang2020needs}. In such multi-agent missions, appropriate trust in robotic collaborators is one of the leading factors influencing HRI performance \cite{khavas2020modeling}. In HRI, factors affecting trust are classified into three categories \cite{hancock2011meta}: 1) \textit{Human-related factors} (i.e., including ability-based factors, characteristics); 2) \textit{Robot-related factors} (i.e., including performance-based factors and attribute-based factors); 3) \textit{Environmental factors} (i.e., including team collaboration, tasking, etc.). Although there is no unified definition of trust in the literature, researchers take a utilitarian approach to defining trust for HRI adopting a trust definition that gives robots practical benefits in developing appropriate behaviors through planning and control \cite{chen2020trust,wang2018trust,kok2020trust}.

Moreover, with the development of the explanation of machine learning, such as explainable AI, although explanation and trust have been intertwined for historical reasons, the utility of machine learning explanations will have more general and consequential \cite{davis2020measure}. Especially the utility of an explanation must be tied to its appropriate trust model.
Although modeling trust has been studied from various perspectives and involves many different factors, it is still challenging to develop a conclusive trust model that incorporates all of these factors. 
Therefore, future research into trust modeling needs to be more focused on developing general trust models based on measures other than the countless factors affecting trust \cite{khavas2020modeling}. Specifically, few works from the literature considered trust in socio-intelligent systems from an AI agent perspective to model trust with agents having similar needs and intrinsic values.

Recently, \cite{yang2021can} proposed a novel trust model termed relative needs entropy (RNE) by combining utility theory and the concept of relative entropy based on the agent's needs hierarchy \cite{yang2020hierarchical,yang2020needs} and innate values. Similar to information entropy, it defines the entropy of needs as the difference or distance of needs distribution between agents in a specific scenario for an individual or group Fig. \ref{fig:needs_trust}. From a statistical perspective, the RNE can be regarded as calculating the similarity of high-dimensional samples from the agent needs vector. A lower RNE value means that the trust level between agents or groups is higher because their needs are well-aligned and there is a low difference (distance) in their needs distributions. Similarly, a higher RNE value will mean that the needs distributions are diverse, and there exists a low trust level between the agent or groups because of the misalignment in their motivations.

For future cognitive robotics research, AI robots learn and adapt to human needs and maintain trust and rapport among robots, which are critical for task efficiency and safety improvement \cite{nourbakhsh2005human}. And we list potential directions as follow:
\begin{itemize}
    \item Adopting suitable formation to perceive and survey environments predicting threats warning human team members.
    \item Reasonable path planning adaptation in various scenarios avoids collision guaranteeing human security and decreasing human working environment interference.
    \item Combining the specific capabilities and needs of robots and humans, calculating sensible strategies to efficiently organize the entire group collaboration and fulfill the corresponding mission.
\end{itemize}

As discussed above, trust is the basis for decision-making in many contexts and the motivation for maintaining long-term relationships based on cooperation and collaboration \cite{cho2015survey}. Although trust is a subjective belief, it is a rational decision based on learning from previous experience to maximize its interest (or utility) and minimize risk by choosing the best compromise between risk and possible utility from cooperation. 
Trust assessment involves various factors, particularly utility and risk analysis under dynamic situations, which rely on context and balance key tradeoffs between conflicting goals to maximize decision effectiveness and task performance.
\begin{figure*}
    \centering
    \includegraphics[width=\textwidth]{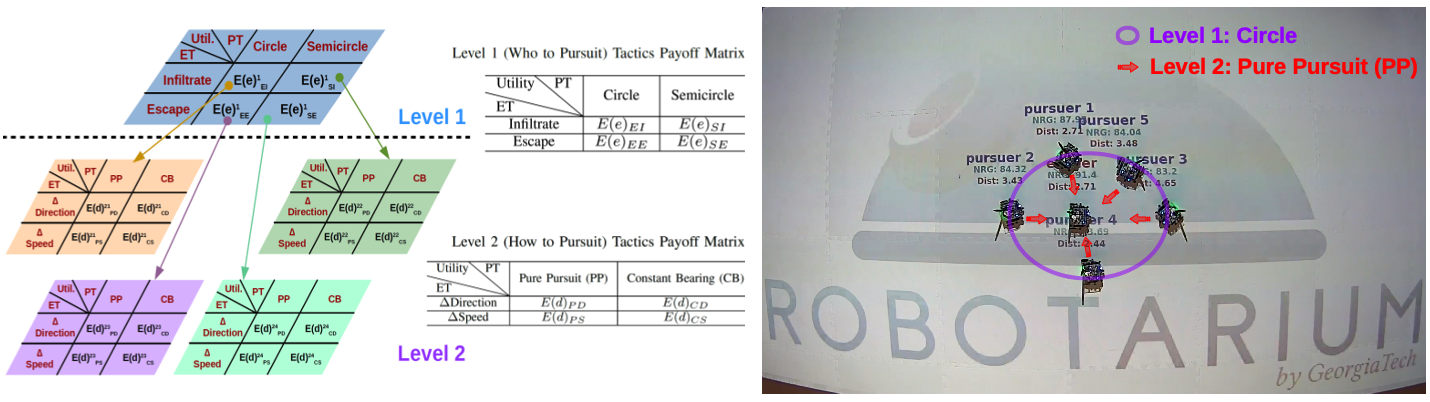}
    \caption{Illustration of five pursuers catching one evader in Pursuit-Evasion game through the game-theoretic utility tree (GUT) with circle formation (level 1) and PP tactic (level 2) in Robotarium \cite{yang2022game}.}
    \label{fig:gut}
\end{figure*}

\section{Human-Robot Interaction}
\label{hri}

As the higher-level intelligent creature, humans represent more complex and diversified needs such as personal security, health, friendship, love, respect, recognition, and so forth. In this regard, humans are highly flexible resources due to the high degree of motion of the physical structure and the high level of intelligence and recognition \cite{toichoa2021emotion}. When we consider humans and robots working as a team, organizing their needs and getting a common ground is a precondition for human-robot collaboration in complex missions and dynamically changing environments. Considering building more stable and reliable relationships between humans and robots, \cite{sanders2011model} depicts an updated model in HRI termed Human-Robot Team Trust (HRTT). It considers the human as a resource with specific abilities that the resource can gain or lose trust with training and realizes the robots as resources with known performance attributes included as needs in the design phase of the collaborative process.

Due to the constantly increasing need for robots to interact, collaborate and assist humans, Human-Robot Interaction (HRI) poses new challenges, such as safety, autonomy, and acceptance issues \cite{zacharaki2020safety}. From a robot needs perspective \cite{yang2020hierarchical}, it first needs to guarantee human security and health, such as avoiding collision with humans, protecting them from risks, etc. Moreover, in the higher level teaming needs, robots should consider human team members' specialty and capability to form corresponding heterogeneous Human-Robot teams adapting specific missions automatically \cite{yang2020needs}. Furthermore, efficient and reliable assistance is essential to the entire mission process in HRI. More importantly, designing an \textit{Interruption Mechanism} can help humans interrupt robots' current actions and re-organize them to fulfill specific emergency tasks or execute some crucial operations manually.

Especially humans also expect robots to provide safety and a stable working environment in missions. \cite{de2013relation} supports the utility of people’s attitudes and anxiety towards robots to explain and predict behavior in human-robot interaction and suggests further investigation of which aspects of robots evoke what type of emotions and how this influences the overall evaluation of robots. From the psychophysiology perspective, \cite{bethel2007survey} combines subjective research methods, such as behavioral and self-report measurements, to achieve a more complete and reliable knowledge of the person’s experience. For example, when people interact with AI agents like robots, which will be affected by mood and social desirability in a specific issue. However, psychophysiological techniques have different challenges in data acquisition and interpretation. Moreover, the monitor quality is influenced by both confounding environmental factors, such as noise or lightning, and by the individual psychological internal state during the evaluation, which might undermine the reliability and the correct interpretation of the gathered data \cite{tiberio2013psychophysiological}.

From another angle, recognizing the potential utility of human affect (e.g., via facial expressions, voice, gestures, etc.) and responding to humans is essential in a human-robot team's collaborative tasks, which would make robots more believable to human observers \cite{breazeal1999build}. \cite{breazeal2004teaching} proposed an architecture including natural language processing, higher-level deliberative functions, and implementing ``joint intention theory''. \cite{scheutz2006utility} demonstrated that expressing affect and responding to human affect with affect expressions can significantly improve team performance in a joint human-robot task. 
\begin{figure*}
\centering
\includegraphics[width=2.05\columnwidth]{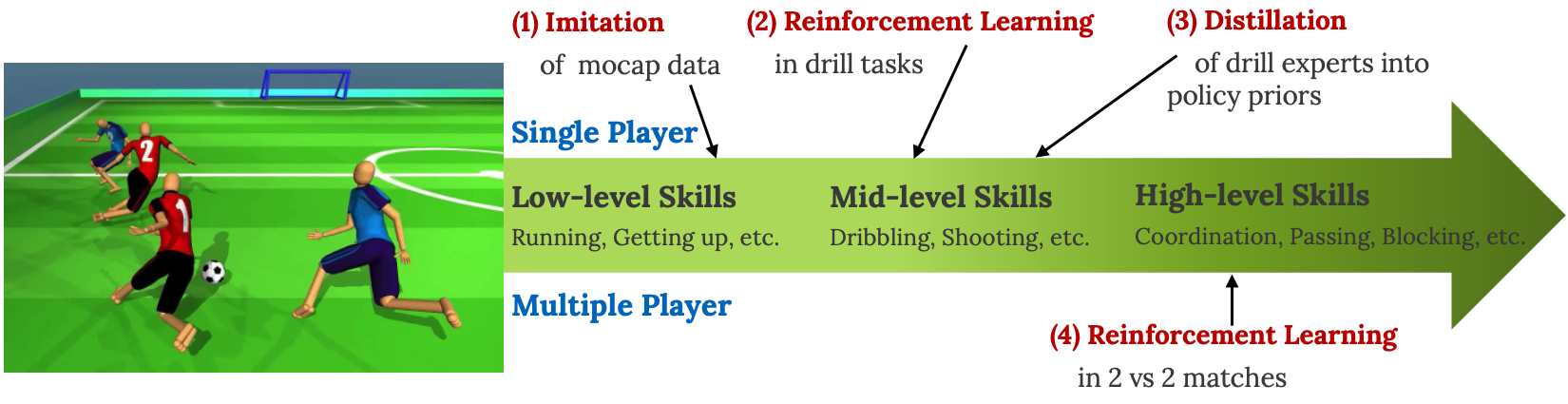}
\caption{End-to-End learning of coordinated 2 vs 2 humanoid football in MuJoCo multi-agent soccer environment \cite{yang2024bayesian}}
\label{fig:humanoid_football}
\end{figure*}

For Multi-agent HRI systems, they require more complex control strategies to coordinate several agents that may be dissimilar to one another (in roles, communication modes etc.), can involve interactions that connect more than two agents at once, and need special attention to address any conflicts that may arise from their interactions \cite{dahiya2023survey}. Specifically, the human would give high-level commands to the robot group or control them individually using a screen-based interface, and robots need to adapt to the operator's actions or optimize some utility function (computational characteristics like task requirements, teaming needs, etc.). 

Moreover, building the shared mental models (SMM) enable team members to draw on their own well-structured common knowledge to select actions that are consistent and coordinated with those of their teammates, which are strongly correlated to team performance \cite{tabrez2020survey}. \cite{nikolaidis2013human} designs the human-robot cross-training inspired by SMM to compute the robot policy aligned with human preference (needs) by iteratively switching roles between humans and robots to learn a shared plan for a collaborative task. Furthermore, considering humans infer the robot’s capabilities and partially adapt to the robot, \cite{nikolaidis2017game} presents a game-theoretic model of human partial adaptation to the robot, where the human response to the robot’s actions by maximizing a reward function that changes stochastically over time, capturing the evolution of their expectations of the robot’s capabilities. Recently, \cite{yang2025edge} introduces the architecture of edge cognitive computing by combining human experts and assisted robots collaborating in the same framework to achieve a seamless remote diagnosis, round-the-clock symptom monitoring, emergency warning, therapy alteration, and advanced assistance. 

In the HRI, {\it adaptability} is a crucial property provided by the SMM, which intrinsically relates to performance and is objectively measured by an adaptable controller as an independent variable to compare alongside other controllers \cite{tabrez2020survey}. Especially in complex, uncertain, and dynamically changing environments, robots need to adapt efficiently to humans' behaviors and strategies caused by their frequently changing needs and motivations. Additionally, \cite{nikolaidis2017human} showed that a robot adapting to the differences in humans’ preferences and needs is positively correlated with trust between them. Moreover, SMM improves trust and reliability between humans and robots by alleviating uncertainty in roles, responsibilities, and capabilities in the interactions. By integrating research regarding trust in automation and describing the dynamics of trust, the role of context, and the influence of display characteristics, \cite{lee2004trust} believes trust involves the experience and knowledge of humans about the needs, motivation, function processing, and performance of robots \cite{hancock2011meta}, such as whether robots' capabilities meet expectations of humans and tasks \cite{kwon2016human}, and minimizing system fault occurrence, system predictability, and transparency \cite{lewis2018role}.

Generally speaking, through formulizing the motivations and needs of humans and robots and describing their complex, diverse, and dynamic relationships in a common ground like trust, utility theory provides a unified model to evaluate the performance of the interaction between humans and AI agents effectively and make a fundamental contribution to safety in the interaction.

\section{Insights and Challenges}
\label{iac}
\begin{figure}
	\centering
    \includegraphics[width=\linewidth]{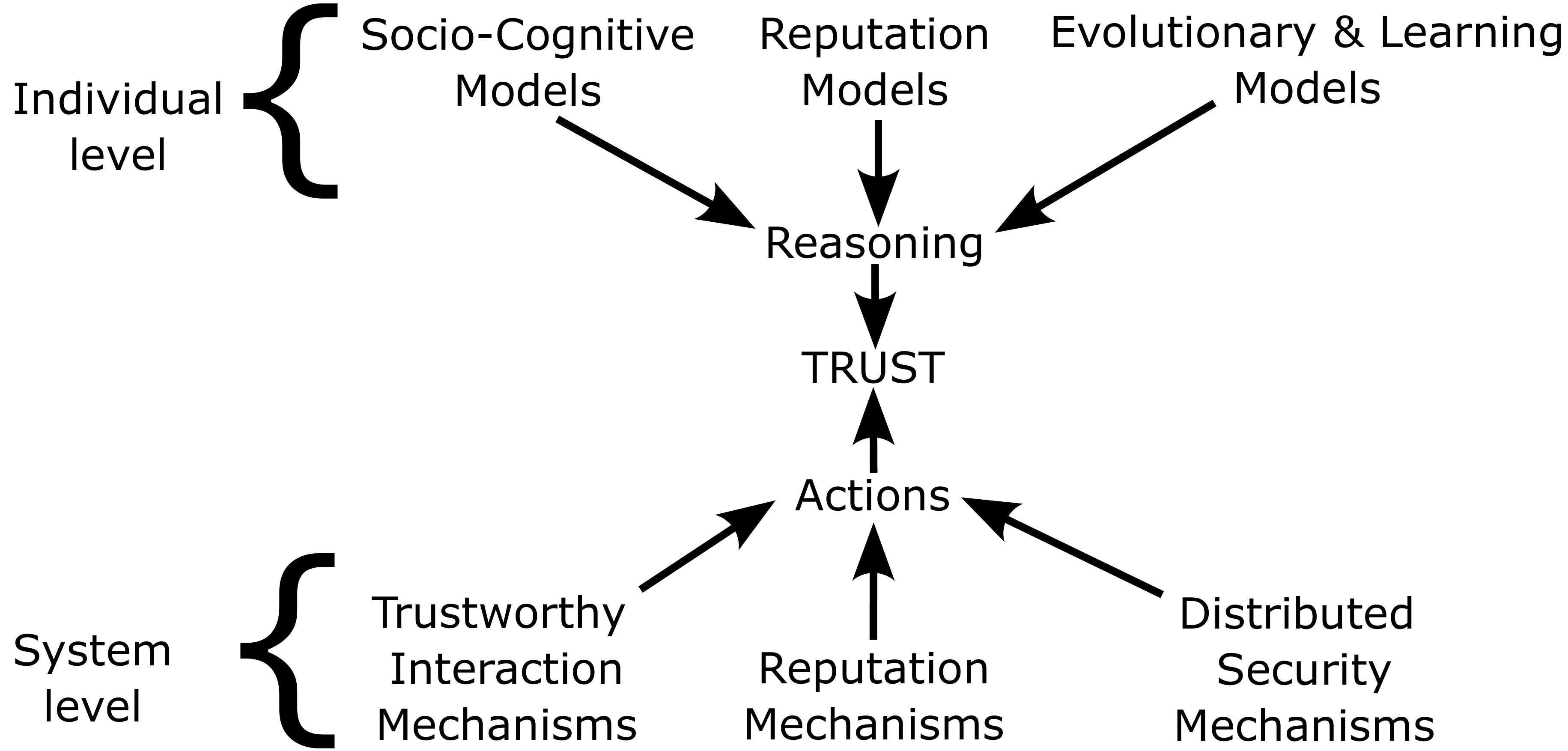}
    \caption{A classification of approaches in trust modeling from individual to multi-agent systems}
    \label{fig:trust}
\end{figure}

As we discussed above, there has been remarkable progress in robotics through utility theory based cognitive modeling, such as intrinsically motivated reinforcement learning systems, diverse values developed and implemented to motivate exploration and learning, and advanced machine learning approaches, like artificial neural networks, reinforcement learning, and Bayesian learning, to build sophisticated value systems. However, several fundamental questions and broad research gaps remain in cognitive robotics: How to build a value system that enables robots to automatically generate motivations in various situations? How can a robot develop an understanding of the structure of its body related to its value system? How can a robot learn the relationships between itself, physical environments, and other agents that inhabit that environment based on its value system?

\subsection{Complex Value Systems}

For biological entities, their value systems represent the brain as having reused many neural elements (Fig. \ref{fig:value_systems}) that evolved originally for sensory pleasures to also mediate higher pleasures \cite{leknes2008common,berridge2014experienced1}, which form our reward utility mechanism for novelty, curiosity, competence, concepts of hunger, thirst, and so on. This mechanism motivates complex behaviours and skills, driving them to improve performance to meet their needs and goals. However, we remain unclear about whether our understanding of motivation and value in natural systems is sufficient to inform the design of complex developmental cognitive systems, especially when modeling them in robots. At this point, we face many grand challenges in cognitive modeling regarding value systems:
\begin{itemize}
    \item How to build a mechanism to share the complexity of different types of motivations and values from natural systems to artificial systems, such as between humans and robots?
    \item How to organize different types of motivations and values to integrate into a single value system for a developmental artificial system?
    \item How to let AI agents explore new values or needs to upgrade their value systems for adapting to dynamic changing environments automatically?
    \item How to explicitly transfer a specific goal from the value system?
    \item How to deal with agents with heterogeneous value systems in their interaction, such as negotiation, consensus, and collaboration?
\end{itemize}

\subsection{Value Systems based Learning}
\begin{figure}
	\centering
    \includegraphics[width=0.9\linewidth]{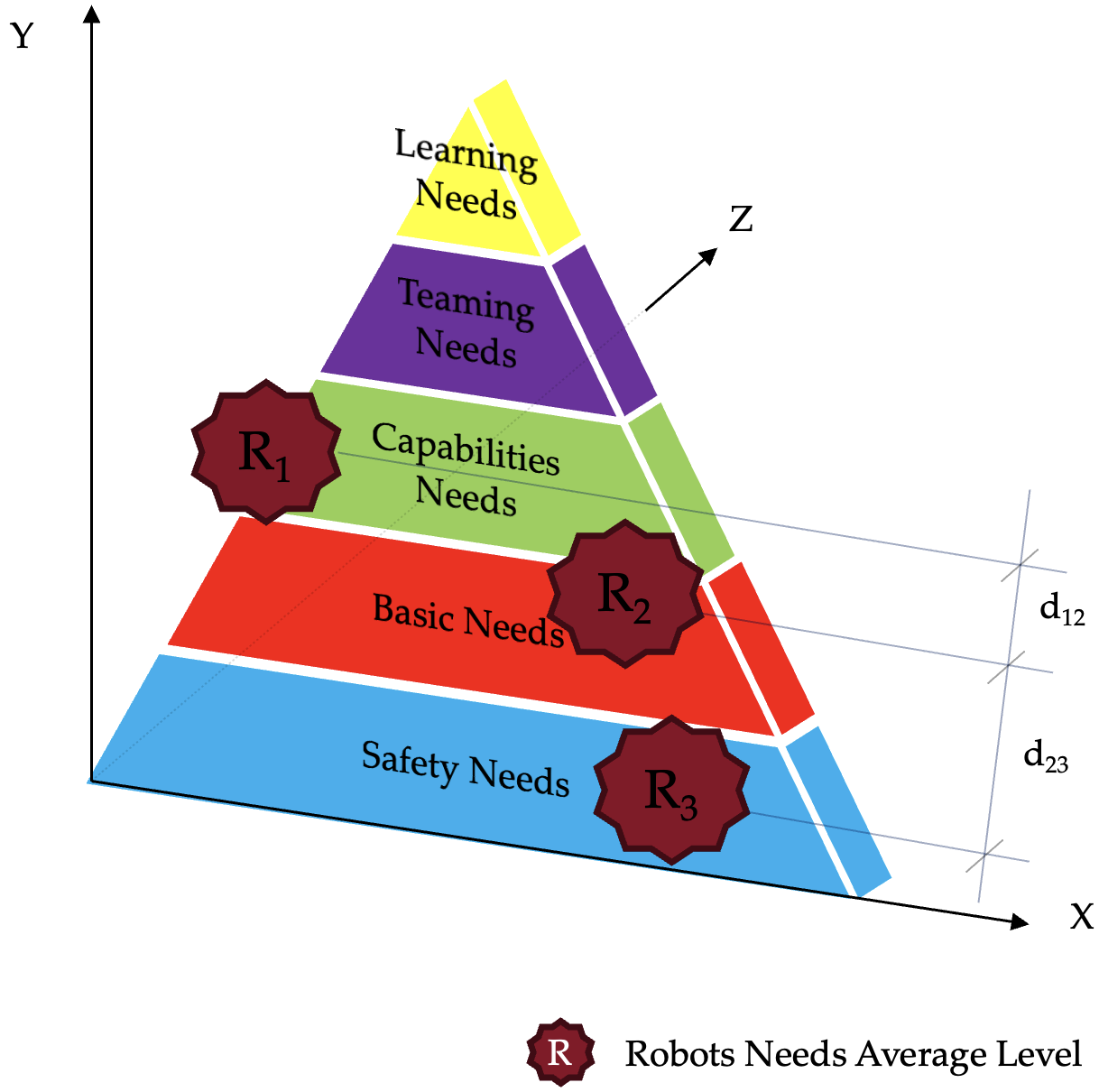}
    \caption{Illustration of The Robots' Trust based on Robot Needs Hierarchy \cite{yang2021can}}
    \label{fig:needs_trust}
\end{figure}

From the learning perspective, the individual learning model can be regarded as constructing models of the other agents, which takes some portion of the observed interaction history as input, and returns a prediction of some property of interest regarding the modeled agent \cite{albrecht2018autonomous}.
At this point, value systems for developmental cognitive robots have been embedded across a range of robotic platforms, including wheeled mobile robots, humanoids, insectoids, and animal robots. However, when we consider multi-degrees of freedom in robot learning, it becomes both time-inefficient and memory-intensive, even with current state-of-the-art approaches. Therefore, developing approaches that enable robots to autonomously shift their attention from one body part to another and to combine skills across different body parts to learn complex, whole-body behaviors and interact with complex environments are much more critical.
Furthermore, another interesting direction is the study of generating behavioural diversity from past experience \cite{benureau2016behavioral} to shape robots such as animals \cite{cully2015robots}, which is complementary to developmental cognitive robotics. It provides ways to transfer knowledge from one task to new and unexpected tasks in a creative way.

Moreover, how to design a general and standard utility mechanism integrating humans and AI agents' needs and learning to efficiently build reliable and stable trust relationships between agents and humans is essential \cite{yang2021can}. Especially a suitable knowledge graph combined with efficient DRL can help agents learn to adopt appropriate behaviors or strategies benefiting the group utilities, adapting to human members' needs, and guaranteeing their development in various situations \cite{yang2024bayesian}. Another interesting direction is behavior manipulation, known as policy elicitation, which refers to problems in human-robot collaboration wherein agents must guide humans toward an optimal policy to fulfill a task through implicit or explicit communication \cite{tabrez2019explanation,chakraborti2017plan}. For example, it is used to achieve effective personalized learning in AI agents' teaching and coaching \cite{leyzberg2018effect}.

\subsection{Artificial Social Systems}
On the other hand, in the development of cognitive modeling in robotics, robots’ cognition will increase, and they will gradually develop self-awareness. They are transforming from traditional roles, like tools and machines, to so-called artificial creatures.
Especially with the tremendous growth in AI technology, Robotics, IoT, and high-speed wireless sensor networks (like 5G), it gradually forms an artificial ecosystem termed {\it artificial social systems} that involves entire AI agents from software entities to hardware devices. How to integrate {\it artificial social systems} into human society and coexist harmoniously is a critical issue for the sustainable development of human beings, such as assistive and healthcare robotics \cite{dragan2012formalizing}, social path planning and navigation \cite{rios2015proxemics}, search and rescue \cite{nourbakhsh2005human,yang2020needs}, smart manufacturing \cite{11135823,ali2025digital}, and autonomous driving \cite{sadigh2018planning}. Especially the future factory is likely to utilize robots for a much broader range of tasks in a much more collaborative manner with humans, which intrinsically requires operation in proximity to humans, raising safety and efficiency issues \cite{tabrez2020survey}. 

Therefore, building a robust, stable, and reliable trust network between humans and AI agents to evaluate agents' performance and status in a common ground is the pre-condition for efficient and safe interaction in their collaboration. Here, utility is the key concept representing the individual agent's innate values, needs, and motivations over states of the world and is described as expected rewards. It measures sensitivity to the impact of actions on trust and long-term cooperation and is efficient enough to allow AI agents like robots to make real-time decisions \cite{kuipers2018can}. For example, self-driving cars might be the first widespread case of trustworthy robots designed to earn trust by demonstrating how well they follow social norms. Additionally, research on joint attention \cite{kaplan2006challenges} between robots, even in human-robot interaction \cite{liu2012generation}, remains a considerable open challenge for modeling social motives such as peer pressure and conformity, especially sharing values or motivations within the group. Moreover, developing an evolutionary value system in cognitive models using utility theory to implement it in cognitive robots, especially for robot swarm interaction \cite{klyne2016intrinsically}, will be a more interesting and challenging direction.

Generally speaking, research on cognitive modeling for autonomous robots is still in its infancy, and there is a long way to go before we have a full-fledged autonomous robot with human-like cognition.

\section{Conclusion}
\label{con}

This survey, from a utility-theoretic perspective, begins with a thorough review of the evolution of cognitive modeling in robotic applications. We discuss cognitive modeling in robotics from BBR and cognitive architectures to the properties of value systems in robots, such as motivations as artificial value systems, and the utility theory based cognitive modeling for generating and updating strategies in robotic interactions. Then, we discussed the applications of biologically inspired motivation systems in single-agent and multi-agent systems. In particular, there has been impressive progress in the design of utility architectures for intrinsic motivation systems implemented in trust evaluation and human-robot interaction. Finally, we analyzed the most important trends of this work to provide a concise representation of the fundamental challenges on which we still need to continue working to deploy these systems in real and lengthy applications.

We believe that cognitive modeling has great potential to accelerate the integration of AI and robotic systems into human society, supporting sustainable human development. This includes alleviating the burden and facilitating the execution of tasks by the most vulnerable sectors of society (such as seniors, children, and the disabled), overcoming the workforce shortage in specific positions related to healthcare, manufacturing, education, etc.

%
%

\bibliographystyle{cas-model2-names}

\bibliography{cas-dc-sample}



\bio{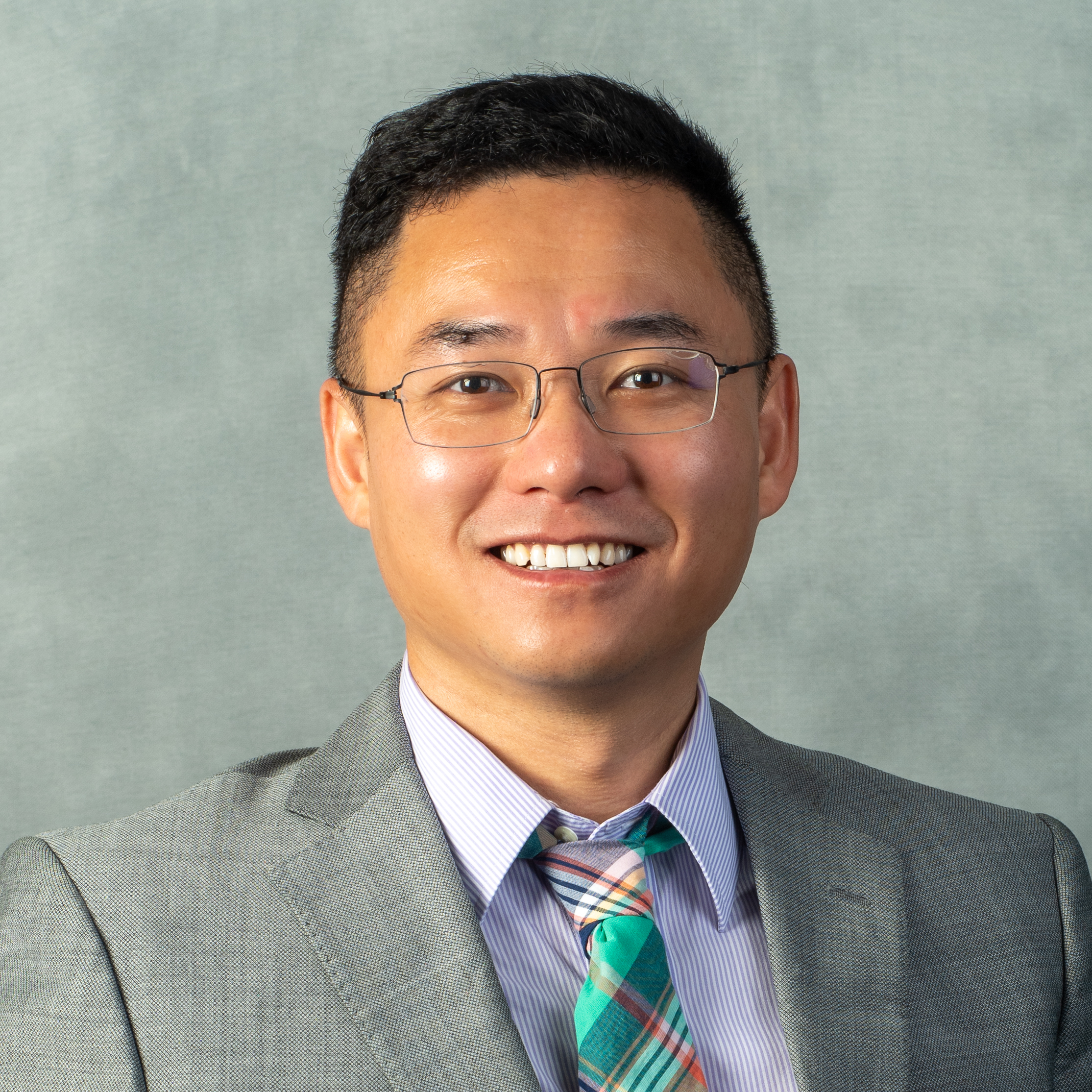}
Qin Yang is an AI and Robotics researcher specializing in Multi-Agent/Robot Systems (MAS/MRS), Artificial Intelligence, Cognitive Modeling, Swarm Intelligence/Robotics, and Human-Robot Interaction (HRI). He is a visiting assistant professor at the University of Tulsa and an assistant professor at Bradley University. Before that, he was a research scientist at Hitachi America Ltd. studying autonomous driving systems. Qin Yang earned his Ph.D. in computer science from the University of Georgia and his MSc in computer science from the Colorado School of Mines. 
He has over 12 years of experience in the Intelligence and embedded Systems field.
\endbio

\end{document}